\documentclass[10pt,journal,compsoc]{IEEEtran}

\usepackage{amsmath}
\usepackage{diagbox}
\usepackage{graphics}
\usepackage{epsfig}
\usepackage{threeparttable}
\usepackage{xspace}

\usepackage{siunitx}

\usepackage{graphicx}
\usepackage{amssymb}
\usepackage{color}
\usepackage[normalem]{ulem}
\usepackage{multirow}
\usepackage{float}
\usepackage{amsfonts}
\usepackage{bm}
\usepackage{array}
\usepackage{colortbl}
\usepackage{rotating}
\usepackage{booktabs}
\usepackage{overpic}
\usepackage{enumitem}
\usepackage{dblfloatfix}

\usepackage[ruled, vlined]{algorithm2e}
\usepackage[textsize=tiny]{todonotes}
\PassOptionsToPackage{table,x11names}{xcolor}
\usepackage{xcolor}
\usepackage{fontawesome}
\usepackage{mdframed}
\usepackage{pifont}
\usepackage{tabularx} 
\usepackage[export]{adjustbox}
\usepackage{subcaption}
\usepackage{newfloat}
\usepackage{listings}
\usepackage{arydshln}
\usepackage{makecell}
\usepackage{caption}
\usepackage{cancel}
\usepackage{wrapfig}
\usepackage{ragged2e}
\usepackage{bbding}
\usepackage{bbm}
\usepackage{mathrsfs}

\definecolor{lightgray}{gray}{.9}
\definecolor{deepgray}{gray}{.8}

\newcolumntype{I}{!{\vrule width 1pt}}
\makeatletter
\newcommand{\thickhline}{%
    \noalign {\ifnum 0=`}\fi \hrule height 1pt
    \futurelet \reserved@a \@xhline
}
\makeatother
\definecolor{mygray}{gray}{.9}

\usepackage[pagebackref=false,breaklinks=true,colorlinks,bookmarks=false]{hyperref}

\definecolor{mydarkdarkred}{RGB}{182, 58, 43}
\definecolor{mydarkdarkred2}{RGB}{180, 92, 67}
\definecolor{mydarkred}{RGB}{245, 220, 215}
\definecolor{mylightred}{RGB}{251, 244, 242}

\definecolor{mydarkdarkblue}{RGB}{17, 24, 129}
\definecolor{mydarkdarkblue2}{RGB}{92, 108, 165}
\definecolor{mydarkblue}{RGB}{211, 218, 234}
\definecolor{mylightblue}{RGB}{241, 244, 250}

\definecolor{mydarkdarkgreen}{RGB}{93, 150, 74}
\definecolor{mydarkgreen}{RGB}{216, 233, 199}
\definecolor{mylightgreen}{RGB}{245, 249, 241}

\definecolor{darkred}{RGB}{139,0,0}

\definecolor{mydarkyellow}{RGB}{216, 214, 196}
\definecolor{mylightyellow}{RGB}{245, 245, 240} 

\definecolor{mygreen}{RGB}{93,173,85}
\definecolor{myorange}{RGB}{233,144,61}
\definecolor{azure}{rgb}{0.0, 0.5, 1.0}
\definecolor{gray}{rgb}{0.3, 0.3, 0.3}
\definecolor{DarkGreen}{RGB}{42,110,63}
\definecolor{DarkBlue}{RGB}{64,101,149}

\definecolor{mylightgray}{RGB}{249, 249, 249}

\definecolor{lightpink}{RGB}{253, 244, 241}
\definecolor{lightblue}{RGB}{245, 255, 255}
\definecolor{lightgreen}{RGB}{255, 255, 241}
\definecolor{lightyellow}{RGB}{255, 252, 240}
\definecolor{oursblue}{HTML}{D7F6FF}
\definecolor{oursrow}{RGB}{255,243,224}

\newcommand{\spm}[1]{\,\raisebox{-0.45ex}{\scalebox{0.62}{\textcolor{gray}{$\pm$#1}}}}

\usepackage[capitalize]{cleveref}
\crefname{section}{Sec.}{Secs.}
\crefname{table}{Tab.}{Tabs.}

\makeatletter
\DeclareRobustCommand\onedot{\futurelet\@let@token\@onedot}
\def\@onedot{\ifx\@let@token.\else.\null\fi\xspace}

\def\eg{\emph{e.g}\onedot} 
\def\ie{\emph{i.e}\onedot}



\begin{document}
\title{Decouple, Purify and Unite:\\
Semantic--Structural Prototype Learning\\
for Federated Medical Segmentation}
\author{Xingyue~Zhao, Wenke~Huang, Linghao~Zhuang, Yanzhou~Su,\\
Zhifeng~Wang, Haoyu~Zhao, Mengfan~Li, Junjun~He,\\
Tao~Tan,~\IEEEmembership{Senior Member,~IEEE,} Dakai~Jin, Le~Lu,~\IEEEmembership{Fellow,~IEEE,}\\
Mang~Ye,~\IEEEmembership{Senior Member,~IEEE,} Qiang~Yang,~\IEEEmembership{Fellow,~IEEE,} and Ming~Feng
\IEEEcompsocitemizethanks{
\IEEEcompsocthanksitem Xingyue Zhao, Wenke Huang, and Linghao Zhuang contributed equally.
\IEEEcompsocthanksitem Corresponding authors: Mang Ye, Qiang Yang, and Ming Feng.
\IEEEcompsocthanksitem X. Zhao and M. Feng are with the Department of Neurosurgery, Peking Union Medical College Hospital, Chinese Academy of Medical Sciences and Peking Union Medical College, Beijing, China (e-mail: fengming@pumch.cn). M. Feng is also with the School of Artificial Intelligence, Beijing University of Posts and Telecommunications, Beijing, China.
\IEEEcompsocthanksitem W. Huang is with the College of Computing and Data Science, Nanyang Technological University, Singapore.
\IEEEcompsocthanksitem L. Zhuang, Z. Wang, J. He, and L. Lu are with Ant Group, China. J. He is also with Shanghai Innovation Institute, Shanghai, China.
\IEEEcompsocthanksitem X. Zhao and L. Zhuang are with The Key Laboratory of Cognition and Decision Intelligence for Complex Systems, Institute of Automation, Chinese Academy of Sciences, Beijing, China.
\IEEEcompsocthanksitem Y. Su, H. Zhao, and D. Jin are with DAMO Academy, Alibaba Group, China.
\IEEEcompsocthanksitem M. Li is with the Department of Medical Engineering, Xijing 986 Hospital, Fourth Military Medical University, Xi'an, China.
\IEEEcompsocthanksitem T. Tan is with the Faculty of Applied Sciences, Macao Polytechnic University, Macao, China.
\IEEEcompsocthanksitem M. Ye is with the School of Computer Science, Wuhan University, Wuhan, China (e-mail: yemang@whu.edu.cn).
\IEEEcompsocthanksitem Q. Yang is with the Department of Data Science and Artificial Intelligence, The Hong Kong Polytechnic University, Hong Kong, China (e-mail: profqiang.yang@polyu.edu.hk).
\IEEEcompsocthanksitem This work was supported in part by the National High Level Hospital Clinical Research Funding under Grant 2025-PUMCH-H-006, in part by the Science and Technology Innovation Program of Xiongan New Area under Grant 2025XAGG0045, in part by the National Natural Science Foundation of China under Grant 623B2080, and in part by the NTU AI-for-X Postdoctoral Fellowship.\vspace{4.5pt}
}}
\markboth{Decouple, Purify and Unite: Semantic--Structural Prototype Learning for Federated Medical Segmentation}%
{Zhao \MakeLowercase{\textit{et al.}}: Decouple, Purify and Unite}

\IEEEtitleabstractindextext{
\begin{abstract}
Federated learning enables medical institutions to train a global model without sharing data, yet feature heterogeneity from diverse scanners or protocols remains challenging. Existing representation-based methods face two limitations: 1) Incomplete Contextual Representation Learning: single-layer or coupled representations overlook multi-level structural cues and entangle regional semantics with boundary details. 2) Layerwise Style and Aggregation Biases: domain-specific style discrepancies across intermediate layers degrade prototypes, while aggregation that overlooks client distribution shifts can further amplify bias. We propose FedBCS+, federated decoupled contextual alignment with style-purified aggregation. We employ Frequency-domain Style Recalibration (FSR) in prototype construction to decouple content-style representations and extract style-purified prototypes. Built upon these purified features, Decoupled Contextual Prototype Alignment (DCPA) explicitly decouples multi-level features into semantic and structural prototypes and aligns regional semantics and fine-grained anatomical structures separately. Style-purified Semantic Prototype Aggregation (S2PA) measures each client's purified prototype divergence from the global consensus and adaptively reweights aggregation toward under-represented clients to reduce consensus bias. On five heterogeneous medical segmentation benchmarks spanning histopathology, MRI, ultrasound, and colonoscopy, FedBCS+ achieves the highest mean Dice among the compared methods. A convergence analysis further characterizes how aggregation and alignment affect the optimization bound.
\end{abstract}
\begin{IEEEkeywords}
Heterogeneous Federated Learning, Medical Image Segmentation, Prototype Learning, Feature Decoupling
\end{IEEEkeywords}}

\newdimen\authorfootnotelimit
\authorfootnotelimit=\dimen\footins
\dimen\footins=180pt
\AddToHookNext{shipout/after}{\global\dimen\footins=\authorfootnotelimit}
\maketitle
\IEEEdisplaynontitleabstractindextext
\IEEEpeerreviewmaketitle
\IEEEraisesectionheading{
\section{Introduction}\label{sec:introduction}}

\begin{figure}[t]
\centering
\includegraphics[width=\columnwidth]{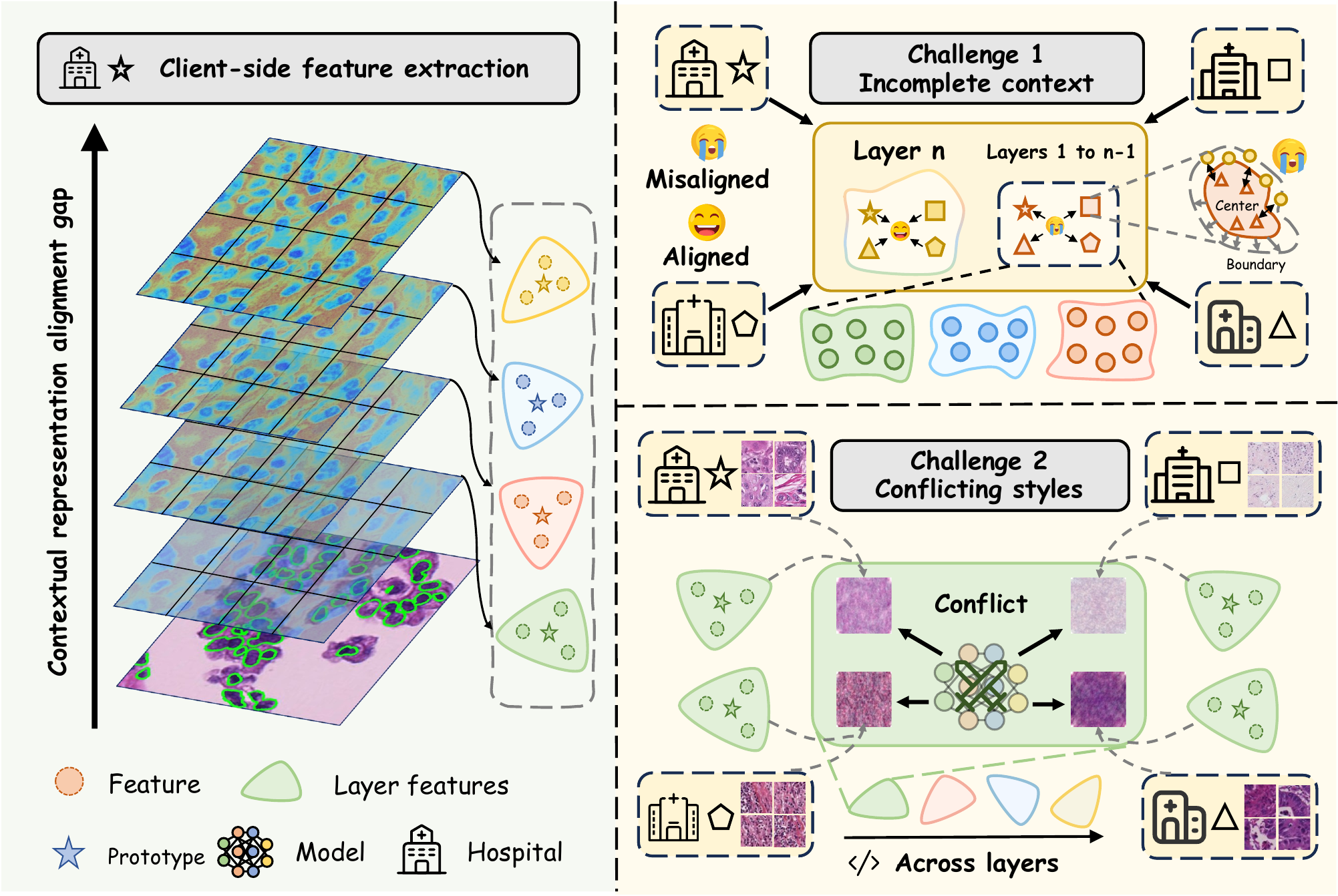}
\caption{\textbf{Two client-side representation challenges in prototype-based federated medical segmentation.}
\textbf{\mbox{Challenge~1}:} Single-layer alignment misses complementary multi-level
context, while prototypes dominated by region-center semantics underrepresent sparse
boundary structures. Coupling these cues in a single prototype further causes
cross-client misalignment.
\textbf{\mbox{Challenge~2}:} Site-specific style discrepancies
propagate across network layers, producing conflicting feature distributions and corrupting
the resulting prototypes. FedBCS+ addresses these issues with DCPA for decoupled
semantic--structural alignment and FSR for style-purified representation learning; the
purified prototypes subsequently guide S2PA aggregation.}
\label{fig:motivation}
\end{figure}

\IEEEPARstart{M}{edical} image segmentation is central to computer-assisted diagnosis {\cite{wang2024cardiovascular}}. The success of current deep learning-based techniques largely relies on massive, centralized datasets. However, clinical data are siloed across institutions, and strict data protection regulations make centralized training impractical. Federated learning (FL) {\cite{FedAvg_AISTATS17}} addresses this by letting multiple medical sites jointly optimize a global model through periodic parameter aggregation, without sharing raw data. Yet in clinical deployments, FL is hampered by data heterogeneity {\cite{FLSurveyandBenchmarkforGenRobFair_TPAMI24, zhao2024morestyle}}. Data from different sites are not independent and identically distributed (non-IID), mainly because of domain shifts caused by different scanners, acquisition protocols, and patient demographics {\cite{jiang2022harmofl, wang2025pfedsam, chen2024think}}. These feature discrepancies pull local models toward their own data distributions, which slows global convergence and degrades segmentation performance. As illustrated in Fig.~\ref{fig:motivation}, applying prototype-based FL to dense medical segmentation raises two client-side representation challenges that motivate our design.

To mitigate data heterogeneity, many existing works regularize local training to improve generalization {\cite{MOON_CVPR21,FedProx_MLSys2020}}. Subsequent works introduce federated prototype learning {\cite{FedProto_AAAI22,FPL_CVPR23}}, and recent efforts further refine these representations via clustering client features {\cite{FedPLVM_NeurIPS24}} or employing Gaussian mixture models {\cite{FedGMKD_NeurIPS24}}. However, these prototype-based methods were designed for classification and face a representational limitation in dense medical image segmentation: \textbf{Incomplete Contextual Representation Learning}. Unlike image-level classification, precise medical segmentation requires modeling both regional pathological semantics (for lesion identification) and fine-grained anatomical structures (for boundary delineation) {\cite{chen2024think, li2025expert, wang2022medical}}. Current FL approaches, largely inheriting paradigms from classification tasks, typically construct prototypes solely from the deepest network bottlenecks. This deep-layer pooling compresses spatial dimensions, forcing both semantic and structural cues into a single, coupled representation. Aligning these coupled features across clients entangles internal semantics with boundary details and causes semantic--structural interference: pursuing a global semantic consensus tends to suppress the multi-level structural cues that dense segmentation requires {\cite{decouple_body_edge_eccv20, sung2024contextrast}}. This raises the first question: \textbf{(a) How to explicitly decouple the feature space to capture more comprehensive context?}

Beyond incomplete contextual representations, existing methods face another limitation spanning both local and server sides of the federated pipeline: \textbf{Layerwise Style and Aggregation Biases}. On the local client side, medical images exhibit significant feature heterogeneity due to diverse scanners and imaging protocols {\cite{jiang2023fair, liu2021feddg}}. Existing methods align final-layer representations {\cite{FPL_CVPR23}}, yet domain-specific style discrepancies remain pronounced in intermediate layers {\cite{huang2017arbitrary, xu2021fourier}}. This persistent layer-wise discrepancy degrades the quality of extracted prototypes before they are even uploaded. At the server, conventional parameter aggregation weights client models by their local sample counts~\cite{FedAvg_AISTATS17}, without using prototype relationships to characterize distribution shifts. Under data skews~\cite{li2022federated, FedNH_AAAI23}, sample counts alone do not capture how style discrepancies affect local representations or how client distributions relate to one another. The resulting aggregation favors dominant domains and compounds the bias introduced during local feature extraction. A reliable global consensus therefore requires aggregation weights that account for the relationships among the uploaded prototypes. Together, the local style corruption and the global aggregation bias raise the second question: \textbf{(b) How to extract style-purified prototypes locally and exploit their correlations globally to rectify the consensus bias?}

\begin{figure*}[t]
\centering
\includegraphics[width=\linewidth]{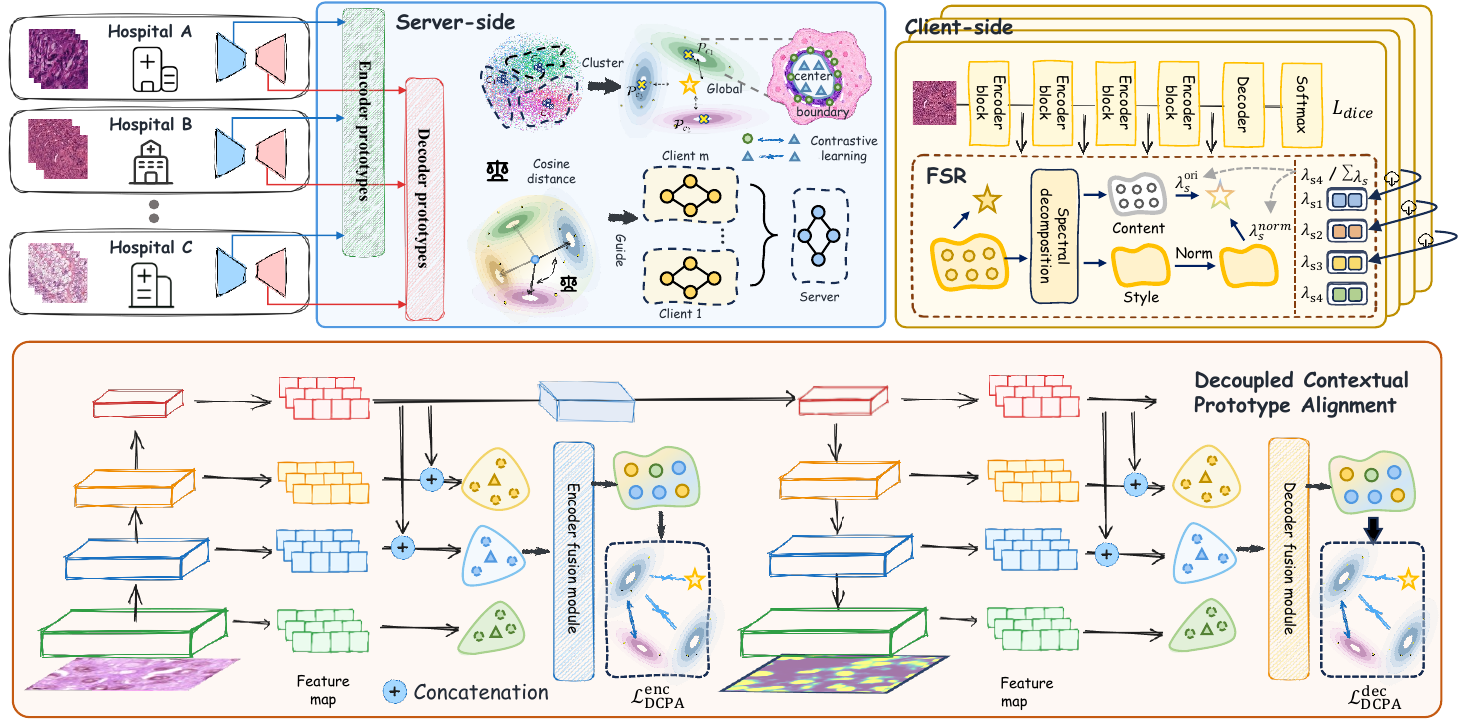}
\caption{\textbf{Overview of FedBCS+.}
On each client, the UNet backbone applies frequency-domain style
recalibration to its deeper encoder stages and extracts multi-level encoder
and decoder features.
DCPA constructs semantic prototypes from background and foreground-center
features and structural prototypes from foreground-boundary features in
each fused stream. The foreground semantic and structural branches are
aligned independently. On the server, S2PA clusters the uploaded prototypes
using FINCH and uses their distances from the global consensus to determine
the parameter aggregation weights.}
\label{fig:framework}
\end{figure*}

To address these challenges, we propose \textbf{FedBCS+}, a federated medical segmentation framework that integrates style recalibration, decoupled prototype alignment, and prototype-guided parameter aggregation. As illustrated in Fig.~\ref{fig:framework}, we first employ Frequency-domain Style Recalibration (FSR) to adaptively modulate the amplitude spectra of intermediate features while preserving their phase information. This suppresses site-specific appearance variations before prototype construction and yields \textit{style-purified features}. From these features, Decoupled Contextual Prototype Alignment (DCPA) integrates multiple encoder and decoder levels to capture both regional semantics and fine-grained anatomical structure. It constructs separate semantic and structural prototypes from region-center and boundary features and aligns the two types independently across clients. This decoupling reduces the interference caused by a shared prototype.

On the server, Style-purified Semantic Prototype Aggregation (S2PA) uses the purified semantic and structural prototypes to guide model aggregation. It measures each client's prototype divergence from the global consensus and converts this information into adaptive parameter aggregation weights. Clients whose representations are under-represented in the consensus receive greater weight, so their distributions are better reflected in the global model. The same prototypes thus serve both local alignment and server-side aggregation.

A preliminary version of this work, FedBCS, was published at AAAI 2026~\cite{FedBCS_AAAI26}. It introduced FSR and Context-aware Dual-level Prototype Alignment (CDPA), which combine style recalibration with hierarchical encoder--decoder feature fusion for federated segmentation. FedBCS+ retains FSR and the multi-level fusion design. The journal version extends FedBCS in four ways:
\begin{itemize}
\item \textbf{Decoupled prototype alignment.} We develop DCPA, extending CDPA from coupled class prototypes to separate semantic and structural prototypes. Independent alignment of the two types reduces the interference between regional semantics and boundary structure and keeps the multi-level feature fusion of CDPA.

\item \textbf{Prototype-guided parameter aggregation.} We introduce S2PA to incorporate representation information into server aggregation. S2PA converts semantic--structural prototype divergence from the global consensus into adaptive model aggregation weights, so that under-represented clients contribute more to the global model.

\item \textbf{Extended convergence analysis.} We extend the conference analysis to account for prototype-guided aggregation and decoupled alignment. The bound distinguishes the contributions of the aggregation-dependent second moment, the positive semantic--structural gradient interaction, and the prototype-anchor perturbation. This analysis identifies sufficient conditions under which FedBCS+ attains a tighter bound than FedBCS.

\item \begingroup
\interlinepenalty=10000
\textbf{Broader empirical validation.} We extend the evaluation from two to five heterogeneous benchmarks by adding BUS, Polyp, and 3D FeTS~2022, which cover ultrasound, colonoscopy, and volumetric MRI. We report five-fold cross-validation results on the four 2D benchmarks and further examine cross-site generalization, component contributions, training dynamics, and representation quality.
\par\endgroup

\end{itemize}

\section{Related Work}
\label{sec:related_work}

\subsection{Federated Learning with Heterogeneous Data}
Federated learning (FL) provides a privacy-preserving framework for collaborative model training without centralizing data~\cite{konevcny2016federated}. FedAvg~\cite{FedAvg_AISTATS17} establishes the basic framework by iteratively aggregating locally trained model parameters. Although effective in homogeneous settings, FedAvg struggles with data heterogeneity, which is common in medical imaging because differences in acquisition protocols, staining procedures, and patient populations lead to substantial distribution shifts across institutions~\cite{wang2022personalizing}. Existing methods address this challenge from several perspectives~\cite{FLSurveyandBenchmarkforGenRobFair_TPAMI24}. Some improve cross-domain performance~\cite{FPL_CVPR23,huang2022learn,GGEUR_CVPR25}, while others emphasize fairness across heterogeneous clients, including group fairness~\cite{ezzeldin2023fairfed}, collaboration fairness~\cite{jiang2023fair}, and performance fairness~\cite{chen2024fair,li2021ditto}. FedBN~\cite{FedBN_ICLR21} retains local batch-normalization statistics to address feature shift, while FCCL+~\cite{huang2023generalizable} supports collaboration across heterogeneous model architectures. Domain generalization methods~\cite{liu2021feddg,zhang2023federated} aim to generalize to unseen distributions. Our work focuses on learning a global segmentation model that generalizes across heterogeneous medical imaging domains.

\subsection{Prototype-based Representation Alignment in Federated Learning}
Many FL methods regularize local training to address data heterogeneity~\cite{wan2024federated,liao2024rethinking,ye2023heterogeneous}; for example, MOON~\cite{MOON_CVPR21} uses contrastive learning to align local and global representations. Prototype-based approaches summarize same-class features as compact mean vectors, reducing the need to share individual feature representations. FedProto~\cite{FedProto_AAAI22} minimizes the distance between local and global class prototypes. However, approaches developed primarily for label skew do not explicitly account for domain shift, limiting their cross-domain generalization. To address domain skew, FPL~\cite{FPL_CVPR23} combines server-side prototype clustering with contrastive learning to align same-class representations, while using a single local prototype per class. FedContrast-GPA~\cite{FedContrast-GPA_MICCAI23} and FedPLVM~\cite{FedPLVM_NeurIPS24} cluster client features to produce multiple prototypes per class, whereas FedGMKD~\cite{FedGMKD_NeurIPS24} constructs prototypes using Gaussian mixture models. These methods nevertheless rely on final-layer encoder features, overlooking both the influence of domain-specific style on prototype quality and the multi-level context required for dense segmentation. Our method constructs style-purified prototypes through frequency-domain recalibration and explicitly decouples semantic and structural representations for fine-grained alignment.

\subsection{Federated Aggregation Strategies}
FedAvg~\cite{FedAvg_AISTATS17} aggregates client parameters with weights proportional to local dataset sizes. Under data heterogeneity, weighting solely by dataset size does not account for differences in client distributions and can yield suboptimal global models~\cite{li2022federated}. Several approaches improve federated optimization and aggregation. FedProx~\cite{FedProx_MLSys2020} adds a proximal term to limit local model drift, and SCAFFOLD~\cite{scaffold2020} introduces control variates to correct drift during local updates. FedNova~\cite{fednova2020} normalizes local updates to address objective inconsistency caused by different numbers of local training steps. Beyond parameter-level corrections, distillation-based methods such as FedDF~\cite{feddf2020} aggregate models by distilling ensemble predictions on a shared unlabeled dataset. For medical FL, FedContrast-GPA~\cite{FedContrast-GPA_MICCAI23} adjusts aggregation weights according to client training dynamics. FedGMKD~\cite{FedGMKD_NeurIPS24} uses discrepancy-aware aggregation to account for distribution differences. These strategies motivate incorporating the semantic content of client representations into aggregation without requiring auxiliary public data. To this end, S2PA measures the divergence between style-purified client prototypes and the global consensus, and converts these measurements into adaptive parameter aggregation weights.

\subsection{Feature Decoupling for Dense Segmentation}
Dense segmentation relies on region-level semantics for target identification and structural cues for precise boundary delineation. Prior work shows that combining these cues in a single representation creates conflicting optimization pressures~\cite{decouple_body_edge_eccv20}. DecoupleNet~\cite{lai2022decouplenet} studies feature decoupling for domain-adaptive segmentation, while Decoupled Semantic Prototypes~\cite{reiss2023decoupled} support learning from diverse annotation types in semi-weakly supervised segmentation. The semantic--structural distinction is particularly relevant to medical imaging: regional semantics identify organs or lesions, while structural cues delineate anatomical boundaries, with both varying substantially across MRI, CT, and histology~\cite{li2025expert, luo2022semi}. In prototype-based FL, coupling these cues within shared prototypes introduces an additional challenge. Aggregating coupled prototypes from clients with different imaging protocols smooths both semantic and structural information, causing their alignment objectives to interfere across communication rounds. DCPA addresses this interference by constructing separate semantic and structural prototypes and aligning them independently.

\newcommand{\FSR}{Frequency-domain Style Recalibration}
\newcommand{\FSRab}{FSR}
\newcommand{\DCPA}{Decoupled Contextual Prototype Alignment}
\newcommand{\DCPAab}{DCPA}
\newcommand{\CDPA}{Context-aware Dual-level Prototype Alignment}
\newcommand{\CDPAab}{CDPA}
\newcommand{\SSPA}{Style-purified Semantic Prototype Aggregation}
\newcommand{\SSPAab}{S2PA}
\newcommand{\oursplus}{FedBCS+}
\newcommand{\oursconf}{FedBCS}

\section{Methodology}\label{sec:method}

We first formulate the problem and then describe \FSRab{}, \DCPAab{}, and \SSPAab{} in Secs.~\ref{sec:fsr}--\ref{sec:s2pa}, followed by a discussion (\cref{sec:discussion}) and a convergence analysis (\cref{sec:convergence}).

\noindent\textbf{Problem Formulation and Notation.}
We consider a standard federated learning setting~\cite{FedAvg_AISTATS17} with $M$ clients indexed by $m$. Each client holds a private dataset $\mathcal{D}_m = \{(x_i, y_i)\}_{i=1}^{N_m}$ of size $N_m$. All clients share a segmentation architecture comprising an encoder $F_{\text{enc}}: \mathcal{X} \rightarrow \mathcal{Z}$, which maps an input image $x$ to feature maps, and a decoder $F_{\text{dec}}: \mathcal{Z} \rightarrow \mathcal{O}$, which produces the segmentation output $\mathbf{o} = F_{\text{dec}}(F_{\text{enc}}(x)) \in \mathbb{R}^{|\mathcal{C}|}$ at each spatial location, where $\mathcal{C}=\{0,1,\dots,|\mathcal{C}|-1\}$ is the set of classes and class $0$ denotes the background. The encoder extracts hierarchical features $\{\mathbf{z}_{\text{enc}}^k\}_{k=1}^{S}$ at $S$ resolution levels, and the decoder produces corresponding features $\{\mathbf{z}_{\text{dec}}^k\}_{k=1}^{S}$. We construct prototypes from a subset of $K\le S$ levels. We focus on \emph{feature heterogeneity}, characterized by $P_m(x|y) \neq P_{m'}(x|y)$: all clients share the same label space, while the conditional feature distribution $P(x|y)$ varies with scanning devices and acquisition protocols. Target sizes, and hence foreground proportions, may also differ across clients (Fig.~\ref{fig:client_het}).
The objective is to collaboratively learn a global model that performs well across all client domains. Appendix~A of the supplementary material provides a consolidated notation table.

\subsection{\FSR}\label{sec:fsr}

FedBCS+ retains the \FSRab{} module introduced in
FedBCS~\cite{FedBCS_AAAI26}, which recalibrates intermediate feature styles
before prototype construction.

\noindent \textbf{Motivation for Style-purified Prototype Construction.}
Existing prototype-based methods assume that same-class features follow similar distributions across clients and therefore construct prototypes by averaging these features. In medical imaging, however, differences in scanning protocols introduce substantial style variations that propagate through network layers and are reflected in the amplitude spectra of intermediate features. Direct averaging retains these discrepancies, so representations of the same anatomical structure can differ markedly across institutions, as shown in Sec.~\ref{sec:fsr_analysis}. Reliable semantic anchors for cross-client alignment therefore require style-purified features. We build on the observation that style and content are largely separable in the frequency domain~\cite{chen2021amplitude, lee2023decompose}, which enables targeted style modulation before prototype construction.

\noindent \textbf{Frequency-domain Recalibration.}
Given an encoder feature map $\mathbf{z}_{\text{enc}} \in \mathbb{R}^{C \times H \times W}$, conventional methods construct class-specific prototypes by direct averaging:
\begin{equation}\small
\setlength\abovedisplayskip{0pt} \setlength\belowdisplayskip{0pt}
p_m^c = \frac{1}{|\mathcal{D}_m^c|} \sum_{x_i \in \mathcal{D}_m^c} F_{\text{enc}}(x_i),
\label{eq:naive_proto}
\end{equation}
where $\mathcal{D}_m^c\subseteq\mathcal{D}_m$ contains the images of client $m$ that include class $c$. These prototypes inherit domain-specific style variations. To address this issue, we first decompose the feature map into amplitude and phase components using a two-dimensional Fourier transform:
\begin{equation}\small
\setlength\abovedisplayskip{0pt} \setlength\belowdisplayskip{0pt}
Z(\alpha, \beta) = \frac{1}{HW} \sum_{h=0}^{H-1} \sum_{w=0}^{W-1} z(h,w) \exp\!\left(-i2\pi \Big(\frac{\alpha h}{H} + \frac{\beta w}{W}\Big)\right).
\label{eq:fft}
\end{equation}
The amplitude spectrum $\chi$ and phase spectrum $\gamma$ are then computed from $Z=Z(\alpha,\beta)$ as
\begin{equation}\small
\setlength\abovedisplayskip{0pt} \setlength\belowdisplayskip{0pt}
\chi(\alpha, \beta) = \sqrt{\operatorname{Re}(Z)^2 + \operatorname{Im}(Z)^2}, \;\;
\gamma(\alpha, \beta) = \arctan \frac{\operatorname{Im}(Z)}{\operatorname{Re}(Z)}.
\label{eq:amp_phase}
\end{equation}
The amplitude spectrum encodes style information, while the phase spectrum preserves semantic content~\cite{chen2021amplitude, lee2023decompose}. We introduce learnable mixing weights to balance normalized and original amplitude information:
\begin{equation}\small
\setlength\abovedisplayskip{0pt} \setlength\belowdisplayskip{0pt}
\lambda_s^{\text{norm}}, \lambda_s^{\text{ori}} = \operatorname{softmax}\!\left(\mathbf{h} / T_s \right),
\label{eq:learnable_weight}
\end{equation}
where $\mathbf{h}$ is a learnable parameter vector and $T_s$ controls the sharpness of the mixing weights. An inverse Fourier transform yields the recalibrated features:
\begin{equation}\small
\setlength\abovedisplayskip{0pt} \setlength\belowdisplayskip{0pt}
\tilde{\mathbf{z}}_{\text{enc}} = \text{IFT}\!\left(\lambda_s^{\text{norm}} \cdot \chi_{\text{norm}} + \lambda_s^{\text{ori}} \cdot \chi, \; \gamma\right),
\label{eq:ifft}
\end{equation}
where $\chi_{\text{norm}}$ denotes the instance-normalized amplitude spectrum. We then apply a channel-wise affine transformation $\tilde{\mathbf{z}}_{\text{enc}} \leftarrow \boldsymbol{\alpha} \odot \tilde{\mathbf{z}}_{\text{enc}} + \boldsymbol{\beta}$, with learnable scale $\boldsymbol{\alpha}$ and bias $\boldsymbol{\beta}$, to restore representational capacity after recalibration. Applying \FSRab{} before averaging gives the style-purified prototypes:
\begin{equation}\small
\setlength\abovedisplayskip{0pt} \setlength\belowdisplayskip{0pt}
\hat{p}_m^c = \frac{1}{|\mathcal{D}_m^c|} \sum_{x_i \in \mathcal{D}_m^c} \tilde{\mathbf{z}}_{\text{enc}}(x_i).
\label{eq:fsr_proto}
\end{equation}
Compared with the prototypes in \cref{eq:naive_proto}, these style-purified prototypes suppress domain-specific style bias while preserving semantic content and are therefore more reliable anchors for cross-domain alignment. These image-level prototypes only illustrate the effect; in \oursplus{}, the prototypes of \cref{sec:dcpa} are computed from the multi-level features of the \FSRab{}-equipped backbone.

\noindent \textbf{Comparison with Related Methods.}
HarmoFL~\cite{jiang2022harmofl} normalizes input amplitudes using shared amplitude statistics and applies weight perturbation to reduce model drift. Instance normalization~\cite{ulyanov2016instance} normalizes feature statistics without adapting the degree of style removal. \FSRab{} instead learns frequency-domain mixing weights, so the balance between content preservation and style removal is adapted during training rather than fixed.

\begin{algorithm}[t]
\SetNoFillComment
\SetArgSty{textnormal}
\small{\KwIn{Communication rounds $R$, local optimization steps $T$, number of clients $M$, private dataset $\mathcal{D}_m$ of client $m$, learning rate $\eta$, hyperparameters $\lambda_{\text{DCPA}}, \lambda_{\text{str}}, \tau, \beta_{\text{ema}}$}}
\BlankLine

\For{$r=1,2,...,R$}{
    \For{$m=1,2,...,M$ \textbf{in parallel}}{
    \tcp{Local Training}
    \For{$t=1,2,...,T$}{
        $\mathbf{z}_{\text{enc}}^k, \mathbf{z}_{\text{dec}}^k \leftarrow$ encoder/decoder features\;
        $\tilde{\mathbf{z}}_{\text{enc}} \leftarrow$ \textbf{\FSRab{}}$(\mathbf{z}_{\text{enc}})$ \tcp*{\cref{eq:ifft}}
        $\hat{\mathbf{z}}_{\text{enc}}^k, \hat{\mathbf{z}}_{\text{dec}}^k \leftarrow$ fusion via \cref{eq:encoder_fusion,eq:decoder_fusion}\;
        $\mathcal{M}_{\text{cen}}, \mathcal{M}_{\text{bdy}} \leftarrow$ erosion via \cref{eq:mask}\;
        $p_{\text{sem}}^{m,c}, p_{\text{str}}^{m,c} \leftarrow$ \cref{eq:dual_proto} for both fused streams\;
        $\mathcal{L}_{\text{local}} = \mathcal{L}_{\text{dice}} + \lambda_{\text{DCPA}}\mathcal{L}_{\text{DCPA}}$ (\cref{eq:total_loss})\;
        $\omega_m \leftarrow \omega_m - \eta \nabla \mathcal{L}_{\text{local}}$\;
    }
    Upload $\omega_m$ and all available $\{p_{\text{sem}}^{m,c}, p_{\text{str}}^{m,c}\}$ to the server\;
    }
    \tcp{Server Aggregation}
    Cluster centroids $\leftarrow$ \textbf{FINCH} on uploaded prototypes; $\bar{p}_{\text{sem}}^c, \bar{p}_{\text{str}}^c \leftarrow$ their means\;
    $d_m \leftarrow$ \textbf{\SSPAab{}} prototype divergence via \cref{eq:proto_dist}\;
    $\tilde w_m^{(r)}, w_m^{(r)} \leftarrow$ weights via \cref{eq:delta_weight,eq:weight_update}\;
    $\omega^{(r+1)} \leftarrow \sum_m w_m^{(r)} \omega_m$ (\cref{eq:weighted_agg})\;
    Broadcast $\omega^{(r+1)}$, the cluster centroids, and $\{\bar{p}_{\text{sem}}^c, \bar{p}_{\text{str}}^c\}$\;
}

\BlankLine
\textbf{\FSRab{}}~($\mathbf{z}_{\text{enc}}$)

{See details in \cref{sec:fsr}}

$\chi, \gamma \leftarrow$ FFT decomposition by \cref{eq:fft,eq:amp_phase}

$\tilde{\mathbf{z}}_{\text{enc}} \leftarrow$ recalibration by \cref{eq:learnable_weight,eq:ifft}

return $\tilde{\mathbf{z}}_{\text{enc}}$

\BlankLine
\textbf{\DCPAab{}}~($\tilde{\mathbf{z}}, y, \{\bar{p}\}$)

{See details in \cref{sec:dcpa}}

$\hat{\mathbf{z}}_{\text{enc}}^k, \hat{\mathbf{z}}_{\text{dec}}^k \leftarrow$ fusion via \cref{eq:encoder_fusion,eq:decoder_fusion}

$p_{\text{sem}}^{m,c}, p_{\text{str}}^{m,c} \leftarrow$ decoupled prototypes via \cref{eq:mask,eq:dual_proto}

$\mathcal{L}_{\text{DCPA}} \leftarrow$ \cref{eq:loss_sem,eq:loss_str,eq:loss_dcpa}

return $\mathcal{L}_{\text{DCPA}}$

\BlankLine
\textbf{\SSPAab{}}~($\{p_{\text{sem}}^{m,c},p_{\text{str}}^{m,c}\},
\{\bar{p}_{\text{sem}}^c,\bar{p}_{\text{str}}^c\}$)

{See details in \cref{sec:s2pa}}

$d_m \leftarrow$ prototype divergence via \cref{eq:proto_dist}

$\tilde w_m,w_m \leftarrow$ weights via \cref{eq:delta_weight,eq:weight_update}

return $\{w_m\}$

\caption{Training procedure for \oursplus{}.}
\label{alg:ours}
\end{algorithm}

\subsection{\DCPA}\label{sec:dcpa}

\textbf{Motivation for Decoupled Representations.}
Existing prototype-based methods~\cite{FedProto_AAAI22, FPL_CVPR23} typically construct and align prototypes using final-layer features. This approach overlooks the hierarchical information distributed across network layers, which captures both fine-grained tissue details and global anatomical context. Our conference version~\cite{FedBCS_AAAI26} introduced \CDPAab{} to fuse multi-level encoder and decoder features into a coupled representation. Although this fusion captures richer context, combining deep semantic cues and shallow structural details in one prototype introduces representational interference: semantic consensus and structural preservation compete within the shared representation, and averaging coupled prototypes tends to smooth anatomical contours. We examine this effect in Sec.~\ref{sec:dcpa_analysis}. To address it, \DCPAab{} retains this hierarchical fusion and extends \CDPAab{} by constructing separate semantic and structural prototypes and aligning them independently.

\noindent \textbf{Multi-level Feature Integration.}
We extract features from $K$ selected levels of the encoder and decoder and use hierarchical fusion to guide shallower features with deeper ones. Let $\{\mathbf{z}_{\text{enc}}^k\}$ denote the selected encoder features. We use the deepest selected layer $k^* = \max\{k\}$ as a semantic anchor and fuse its features with those of each shallower layer:
\begin{equation}\small
\setlength\abovedisplayskip{0pt} \setlength\belowdisplayskip{0pt}
\hat{\mathbf{z}}_{\text{enc}}^k = \phi_{\text{enc}}^k\!\left(\left[\mathbf{z}_{\text{enc}}^k; \; \text{Up}(\mathbf{z}_{\text{enc}}^{k^*})\right]\right), \quad k < k^*,
\label{eq:encoder_fusion}
\end{equation}
where $[\cdot;\cdot]$ denotes channel-wise concatenation, $\text{Up}(\cdot)$ denotes nearest-neighbor upsampling, and $\phi_{\text{enc}}^k$ is a $1 {\times} 1$ convolution. This fusion combines high-level semantic guidance with local detail. We apply the same strategy to the decoder:
\begin{equation}\small
\setlength\abovedisplayskip{0pt} \setlength\belowdisplayskip{0pt}
\hat{\mathbf{z}}_{\text{dec}}^k = \phi_{\text{dec}}^k\!\left(\left[\mathbf{z}_{\text{dec}}^k; \; \text{Up}(\mathbf{z}_{\text{dec}}^{k^*})\right]\right), \quad k < k^*.
\label{eq:decoder_fusion}
\end{equation}
The fused features are aggregated across selected layers via spatial alignment and concatenation: $\mathbf{z}_{\text{enc}}^{\text{fused}} = \bigoplus_{k<k^*} \text{Resize}(\hat{\mathbf{z}}_{\text{enc}}^k)$ and $\mathbf{z}_{\text{dec}}^{\text{fused}} = \bigoplus_{k<k^*} \text{Resize}(\hat{\mathbf{z}}_{\text{dec}}^k)$, where $\bigoplus$ denotes channel-wise concatenation and $\text{Resize}(\cdot)$ aligns spatial resolutions. The anchor layer $k^*$ enters the fusion only as semantic guidance in \cref{eq:encoder_fusion,eq:decoder_fusion} and contributes no separate prototype. We denote these two fused streams by $\mathcal{R}=\{\mathrm{enc},\mathrm{dec}\}$, indexed by $e$. The prototype construction below is applied independently to each stream, with the stream index omitted for clarity; the resulting losses are averaged over $\mathcal{R}$.

\noindent \textbf{Spatial Decoupling into Semantic and Structural Prototypes.}
In medical image segmentation, the interior of an organ encodes its semantic identity (\eg, tissue type), while its boundary provides structural cues for precise delineation. Given a ground-truth label map $y$, we extract the foreground mask $\mathcal{M}_{\text{fg}} = \mathbbm{1}[y > 0]$ and apply morphological erosion:
\begin{equation}\small
\setlength\abovedisplayskip{0pt} \setlength\belowdisplayskip{0pt}
\mathcal{M}_{\text{cen}} = \mathcal{E}_\delta(\mathcal{M}_{\text{fg}}), \quad \mathcal{M}_{\text{bdy}} = \mathcal{M}_{\text{fg}} - \mathcal{M}_{\text{cen}},
\label{eq:mask}
\end{equation}
where $\mathcal{E}_\delta(\cdot)$ denotes morphological erosion with radius $\delta$, implemented efficiently as $-\text{MaxPool}(-\cdot)$ with kernel size $2\delta{+}1$. $\mathcal{M}_{\text{cen}}$ identifies the foreground interior, and $\mathcal{M}_{\text{bdy}}$ captures the foreground boundary strip of width $\delta$ pixels. Let $\mathcal{C}_{\text{fg}}=\mathcal{C}\setminus\{0\}$ denote the foreground classes. The background is retained as semantic context without erosion, giving $\Omega_0^{\text{sem}}=\{v\mid y_v=0\}$. For each $c\in\mathcal{C}_{\text{fg}}$, we define $\Omega_c^{\text{sem}}=\{v\mid y_v=c\wedge\mathcal{M}_{\text{cen}}(v)=1\}$ and $\Omega_c^{\text{str}}=\{v\mid y_v=c\wedge\mathcal{M}_{\text{bdy}}(v)=1\}$. We further define the valid class sets $\mathcal{C}_{\text{sem}}=\{c\in\mathcal{C}\mid |\Omega_c^{\text{sem}}|>0\}$ and $\mathcal{C}_{\text{str}}=\{c\in\mathcal{C}_{\text{fg}}\mid |\Omega_c^{\text{str}}|>0\}$. The corresponding prototypes are:
\begin{equation}\small
\setlength\abovedisplayskip{0pt} \setlength\belowdisplayskip{0pt}
\begin{aligned}
p_{\text{sem}}^{m,c}
&= \frac{1}{|\Omega_c^{\text{sem}}|}
\sum_{v \in \Omega_c^{\text{sem}}}\mathbf{z}_v,
&& c\in\mathcal{C}_{\text{sem}},\\[-2pt]
p_{\text{str}}^{m,c}
&= \frac{1}{|\Omega_c^{\text{str}}|}
\sum_{v \in \Omega_c^{\text{str}}}\mathbf{z}_v,
&& c\in\mathcal{C}_{\text{str}}.
\end{aligned}
\label{eq:dual_proto}
\end{equation}
The background therefore contributes one semantic prototype, while each valid foreground class contributes a semantic prototype from its center region and a structural prototype from its boundary strip. We construct prototypes only for nonempty regions: if erosion removes a foreground center or a class is absent from the current batch, we omit the corresponding prototype and loss term.

\noindent \textbf{Decoupled Contrastive Alignment.}
Given the cluster centroids and consensus prototypes $\{\bar{p}_{\text{sem}}^c, \bar{p}_{\text{str}}^c\}$ received from the server (\cref{sec:s2pa}), we define a separate alignment loss for each prototype type. Let $\mathcal{C}_{\text{sem}}^{\text{fg}}=\mathcal{C}_{\text{sem}}\cap\mathcal{C}_{\text{fg}}$ denote the valid foreground semantic classes. For \textit{semantic alignment}, we combine InfoNCE with mean-alignment regularization:
\begin{equation}\small
\setlength\abovedisplayskip{0pt} \setlength\belowdisplayskip{0pt}
\begin{aligned}
\mathcal{L}_{\text{sem}}
={}& \frac{1}{|\mathcal{C}_{\text{sem}}^{\text{fg}}|}
\sum_{c \in \mathcal{C}_{\text{sem}}^{\text{fg}}}
\frac{1}{|\Omega_c^{\text{sem}}|}
\sum_{v \in \Omega_c^{\text{sem}}} \Bigg[
\\[-2pt]
&-\log
\frac{\sum_{q \in \mathcal{Q}_{\text{sem}}^{+,c}}
\exp(\text{sim}(\mathbf{z}_v,q)/\tau)}
{\sum_{q \in \mathcal{Q}_{\text{sem}}}
\exp(\text{sim}(\mathbf{z}_v,q)/\tau)}
\\[-2pt]
&+\left\|\mathbf{z}_v-\bar{p}_{\text{sem}}^c\right\|_2^2
\Bigg],
\end{aligned}
\label{eq:loss_sem}
\end{equation}
where $\mathcal{Q}_{\text{sem}}^{+,c}$ contains the semantic cluster centroids of foreground class $c$, $\mathcal{Q}_{\text{sem}}$ contains the semantic centroids of all classes, including the background centroids as negative anchors, and $\text{sim}(\cdot,\cdot)$ denotes cosine similarity. The contrastive term encourages inter-class separability, while the MSE term pulls center-region features toward their class prototype to encourage intra-class compactness. We set $\mathcal{L}_{\text{sem}}=0$ when $\mathcal{C}_{\text{sem}}^{\text{fg}}$ is empty. For \textit{structural alignment}, we design a boundary-aware contrastive loss:
\begin{equation}\small
\setlength\abovedisplayskip{0pt} \setlength\belowdisplayskip{0pt}
\begin{aligned}
\mathcal{L}_{\text{str}}
={}&-\frac{1}{|\mathcal{C}_{\text{str}}|}
\sum_{c \in \mathcal{C}_{\text{str}}}
\frac{1}{|\Omega_c^{\text{str}}|}
\sum_{v \in \Omega_c^{\text{str}}}
\\[-2pt]
&\log
\frac{\sum_{q \in \mathcal{Q}_{\text{str}}^{+,c}}
\exp(\text{sim}(\mathbf{z}_v,q)/\tau)}
{\sum_{q \in \mathcal{Q}_{\text{str}}^c}
\exp(\text{sim}(\mathbf{z}_v,q)/\tau)},
\end{aligned}
\label{eq:loss_str}
\end{equation}
where $\mathcal{Q}_{\text{str}}^{+,c}$ contains the structural cluster centroids of class $c$, and $\mathcal{Q}_{\text{str}}^c = \mathcal{Q}_{\text{str}}^{+,c} \cup \mathcal{Q}_{\text{str}}^{-,c}$ adds the negative set. The negative set $\mathcal{Q}_{\text{str}}^{-,c}$ comprises the semantic centroids of class $c$ and the centroids of all other classes. This loss separates boundary features from both same-class center features and other-class features, preserving the semantic--structural distinction and inter-class separability. We set $\mathcal{L}_{\text{str}}=0$ when $\mathcal{C}_{\text{str}}$ is empty.
Restoring the stream index omitted above, the total \DCPAab{} loss averages
the two independently aligned components over the fused streams:
\begin{equation}\small
\setlength\abovedisplayskip{0pt} \setlength\belowdisplayskip{0pt}
\mathcal{L}_{\text{DCPA}}
=
\frac{1}{|\mathcal{R}|}
\sum_{e\in\mathcal{R}}
\left(
\mathcal{L}_{\text{sem}}^{e}
+
\lambda_{\text{str}}\mathcal{L}_{\text{str}}^{e}
\right).
\label{eq:loss_dcpa}
\end{equation}

\subsection{\SSPA}\label{sec:s2pa}

\textbf{Motivation for Prototype-guided Aggregation.}
FedBCS~\cite{FedBCS_AAAI26} uses FedAvg~\cite{FedAvg_AISTATS17} for server-side parameter aggregation, with weights proportional to each client's sample count. \SSPAab{} instead derives the aggregation weights from representation differences across clients. Style-purified semantic and structural prototypes provide complementary measures of each client's divergence from the global consensus, and this information can therefore guide aggregation.

\noindent \textbf{Prototype Divergence Computation.}
At the end of local training in each communication round, client $m$ uploads its style-purified semantic and structural prototypes along with its model parameters. The server applies FINCH clustering~\cite{sarfraz2019efficient} separately to prototypes from the same stream, type, and class, and then averages the resulting centroids to obtain the corresponding global consensus prototype. Both are broadcast: the centroids serve as positive and negative anchors in \cref{eq:loss_sem,eq:loss_str}, and the consensus prototypes enter the MSE term of \cref{eq:loss_sem} and the divergence in \cref{eq:proto_dist}. Let $\mathcal{I}_m$ denote the set of prototypes matched between client $m$ and the global consensus, where each index $i\in\mathcal{I}_m$ identifies a fused stream, a prototype type (semantic or structural), and a class, with client prototype $p_i^m$ and consensus prototype $\bar p_i$. We compute the prototype divergence of client $m$ as
\begin{equation}\small
\setlength\abovedisplayskip{0pt} \setlength\belowdisplayskip{0pt}
d_m =
\frac{1}{|\mathcal{I}_m|}
\sum_{i\in\mathcal{I}_m}
\left(
1-
\frac{
p_i^{m}\cdot\bar p_i
}{
\|p_i^{m}\|_2\,
\|\bar p_i\|_2
}
\right),
\label{eq:proto_dist}
\end{equation}
which averages the cosine distances over all matched semantic and structural prototypes from both fused streams.

\noindent \textbf{Adaptive Weight Assignment.}
We convert these prototype divergences into aggregation weights using a momentum update for stability:
\begin{equation}\small
\setlength\abovedisplayskip{0pt} \setlength\belowdisplayskip{0pt}
\Delta w_m^{(r)} = (1 - \beta_{\text{ema}}) \cdot \Delta w_m^{(r-1)} + \beta_{\text{ema}} \cdot \frac{d_m}{\sum_{j \in \mathcal{S}} d_j},
\label{eq:delta_weight}
\end{equation}
\begin{equation}\small
\setlength\abovedisplayskip{0pt} \setlength\belowdisplayskip{0pt}
\begin{aligned}
\tilde w_m^{(r)}
&= \tilde w_m^{(r-1)}+\Delta w_m^{(r)},\\
w_m^{(r)}
&= \frac{\tilde w_m^{(r)}}{\sum_{j\in\mathcal{S}}\tilde w_j^{(r)}},
\end{aligned}
\label{eq:weight_update}
\end{equation}
where $\beta_{\text{ema}}$ is the momentum coefficient, $\mathcal{S}$ is the set of participating clients, $\Delta w_m^{(0)}=0$, and $\tilde w_m^{(0)}=1/M$. Larger prototype divergences contribute larger increments to the accumulated aggregation scores. Clients whose representations are under-represented in the global consensus therefore receive larger aggregation weights. We update the global model as
\begin{equation}\small
\setlength\abovedisplayskip{0pt} \setlength\belowdisplayskip{0pt}
\omega^{(r+1)} = \sum_{m \in \mathcal{S}} w_m^{(r)} \cdot \omega_m^{(r)}.
\label{eq:weighted_agg}
\end{equation}
Whereas FedAvg uses $w_m = N_m / \sum_j N_j$, \SSPAab{} sets each client's weight from its prototype divergence.

\noindent \textbf{Comparison with Related Methods.}
FedContrast-GPA~\cite{FedContrast-GPA_MICCAI23} adjusts aggregation weights according to client training dynamics, and FedGMKD~\cite{FedGMKD_NeurIPS24} uses discrepancy-aware aggregation. \SSPAab{} instead derives the weights from the divergence of style-purified semantic and structural prototypes from the global consensus. Because \FSRab{} suppresses domain-specific style before the prototypes are computed, this divergence is less affected by appearance differences across sites.

\subsection{Discussion}
\label{sec:discussion}

\noindent\textbf{Overall Framework.}
Fig.~\ref{fig:framework} and Algorithm~\ref{alg:ours} summarize \oursplus{}, which integrates prototype construction, feature alignment, and server aggregation. First, \FSRab{} (\cref{sec:fsr}) recalibrates intermediate features in the frequency domain to suppress style bias before prototype construction. Next, \DCPAab{} (\cref{sec:dcpa}) constructs separate semantic and structural prototypes from the fused multi-level features and aligns them independently. Finally, \SSPAab{} (\cref{sec:s2pa}) uses their divergence from the global consensus to adaptively weight client models and correct aggregation bias. The local training objective is
\begin{equation}\small
\setlength\abovedisplayskip{0pt} \setlength\belowdisplayskip{0pt}
\mathcal{L}_{\text{local}}
=
\mathcal{L}_{\text{dice}}
+
\lambda_{\text{DCPA}}\mathcal{L}_{\text{DCPA}}.
\label{eq:total_loss}
\end{equation}

\noindent\textbf{Comparison with the Conference Version.}
Both \oursconf{} and \oursplus{} use \FSRab{} and multi-level encoder--decoder feature fusion. \oursconf{}~\cite{FedBCS_AAAI26} uses \CDPAab{} to construct a coupled prototype per class and employs FedAvg for server aggregation. Its local objective and aggregation rule are
\begin{equation}\small
\setlength\abovedisplayskip{0pt} \setlength\belowdisplayskip{0pt}
\mathcal{L}_{\text{local}}^{\text{conf}} = \mathcal{L}_{\text{dice}} + \mathcal{L}_{\text{CDPA}}, \quad \omega^{(r+1)} = \frac{\sum_m N_m \cdot \omega_m}{\sum_m N_m}.
\label{eq:conf_loss}
\end{equation}
\oursplus{} makes two changes to the training procedure. First, $\mathcal{L}_{\text{DCPA}}$ replaces $\mathcal{L}_{\text{CDPA}}$, so semantic and structural prototypes are aligned separately rather than through one coupled prototype. Second, \SSPAab{} determines the parameter aggregation weights from prototype information. The resulting objective is \cref{eq:total_loss}, and the aggregation rule is \cref{eq:weighted_agg}.

\noindent\textbf{Communication and Computational Complexity.}
Each fused stream contributes at most $|\mathcal{C}|$ semantic prototypes (one for the background and $|\mathcal{C}|{-}1$ for the foreground classes) and $|\mathcal{C}|{-}1$ foreground structural prototypes. The per-client prototype payload is therefore $\mathcal{O}(|\mathcal{R}|(2|\mathcal{C}|{-}1)d)$, where $|\mathcal{R}|=2$ denotes the fused encoder and decoder streams and $d$ is the prototype dimension. Binary segmentation requires at most six prototypes; prototypes for missing classes or empty eroded regions are omitted.
For local computation, \DCPAab{} adds morphological erosion, implemented by a single max-pooling pass with complexity $\mathcal{O}(H {\times} W)$, and per-pixel InfoNCE against the cluster centroids. Both operations are lightweight relative to a standard forward and backward pass. \FSRab{} adds a two-dimensional FFT/IFFT pair with complexity $\mathcal{O}(C {\times} H {\times} W {\times} \log(HW))$ for $C$ channels. On the server, \SSPAab{} computes the prototype divergences with complexity $\mathcal{O}(M|\mathcal{R}|(2|\mathcal{C}|{-}1)d)$. \Cref{sec:overhead} compares communication and computation costs across methods.

\subsection{Convergence Analysis}
\label{sec:convergence}
We analyze an idealized aggregated-direction surrogate of FedBCS+ to
characterize the contributions of the aggregation-dependent second moment,
the semantic--structural gradient interaction, and the prototype-anchor
perturbation to the optimization bound. Building on our conference
analysis~\cite{FedBCS_AAAI26}, we present the main results here and provide
detailed derivations in Appendices~B--F of the supplementary material. Let $R$ denote the
number of communication rounds and $T$ the number of surrogate optimization
steps per round. For client
$m$, define $\mathcal{F}_m(\Theta)$ as the local objective over the model
parameters $\Theta$, consisting of
$\mathcal{L}_{\mathrm{dice}}$ and
$\lambda_{\mathrm{DCPA}}
(\mathcal{L}_{\mathrm{sem}}+\lambda_{\mathrm{str}}\mathcal{L}_{\mathrm{str}})$.
For an aggregation vector
$w=\{w_m\}_{m=1}^{M}$ satisfying $w_m\geq 0$ and
$\sum_{m=1}^{M}w_m=1$, the global objective is
\begin{equation}
\mathcal{F}_{w}(\Theta)=\sum_{m=1}^M w_m\mathcal{F}_m(\Theta).
\label{eq:weighted_objective}
\end{equation}
The conference FedBCS bound is recovered by setting
$w_m=N_m/\sum_jN_j$ and using the coupled \CDPAab{} loss. In contrast,
\oursplus{} uses \SSPAab{} weights and the decoupled \DCPAab{} objective.

\noindent\textbf{Assumptions.}
At the beginning of round $r$, the aggregation vector $w^{(r)}$
is determined from the available server history and is held
fixed within that round. We analyze the surrogate trajectory
\begin{equation}
	\Theta_{r,\ell+1}
	=
	\Theta_{r,\ell}
	-\eta\widetilde g_{r,\ell},
	\qquad
	\ell=0,\ldots,T-1,
\end{equation}
where
\[
\widetilde g_{r,\ell}
=
g_{r,\ell}
+
b_{r,\ell}.
\]
Here, $g_{r,\ell}$ denotes an ideal stochastic direction for the
weighted objective $\mathcal{F}_w$, while $b_{r,\ell}$ denotes the
optimization perturbation associated with finite and clustered
global prototype anchors. The analysis relies on the following conditions:
(1) every $\mathcal{F}_m$ is $L_{\mathrm{sm}}$-smooth; (2) given the history $\mathcal{H}_{r,\ell}$ before step $\ell$ of round $r$, the ideal direction is conditionally unbiased,
\begin{equation}
	\mathbb{E}
	\left[
	g_{r,\ell}\mid\mathcal H_{r,\ell}
	\right]
	=
	\nabla \mathcal{F}_w(\Theta_{r,\ell});
\end{equation}
(3) its conditional second moment satisfies
\begin{equation}
	\mathbb{E}
	\left[
	\|g_{r,\ell}\|^2
	\mid\mathcal H_{r,\ell}
	\right]
	\leq
	\|\nabla \mathcal{F}_w(\Theta_{r,\ell})\|^2
	+
		\bar{\sigma}_{w,\rho_+}^2;
\end{equation}
and (4) the additional one-step contribution of the
prototype-anchor perturbation is bounded as
\begin{align}
	&\mathbb{E}\Bigg[
	-\eta
	\left\langle
	\nabla \mathcal{F}_w(\Theta_{r,\ell}),
	b_{r,\ell}
	\right\rangle
	\nonumber\\
	&\quad+
	\frac{L_{\mathrm{sm}}\eta^2}{2}
	\left(
	\|g_{r,\ell}+b_{r,\ell}\|^2
	-
	\|g_{r,\ell}\|^2
	\right)
	\Bigm|
	\mathcal H_{r,\ell}
	\Bigg]
	\leq
	\eta C_{\mathrm{pa}},
\end{align}
where $C_{\mathrm{pa}}\geq 0$ is an abstract
prototype-anchor perturbation constant. For interpretation, we write
\begin{equation}
	\bar{\sigma}_{w,\rho_+}^2
	=
	\sigma_w^2
	+
	2\lambda_{\mathrm{DCPA}}^2
	\lambda_{\mathrm{str}}\rho_+,
\end{equation}
where $\sigma_w^2$ absorbs the remaining stochastic,
aggregation-dependent, and surrogate-drift contributions, and
\begin{equation}
	\rho_+
	=
	\mathbb{E}
	\left[
	\max
	\left(
	0,
	\left\langle
	\nabla \mathcal{L}_{\mathrm{sem}},
	\nabla \mathcal{L}_{\mathrm{str}}
	\right\rangle
	\right)
	\right].
\end{equation}
The positive part arises from bounding the cross term:
\[
\begin{aligned}
&2\lambda_{\mathrm{DCPA}}^2\lambda_{\mathrm{str}}
\left\langle
\nabla\mathcal{L}_{\mathrm{sem}},
\nabla\mathcal{L}_{\mathrm{str}}
\right\rangle\\
&\quad\leq
2\lambda_{\mathrm{DCPA}}^2\lambda_{\mathrm{str}}
\max\!\left(
0,
\left\langle
\nabla\mathcal{L}_{\mathrm{sem}},
\nabla\mathcal{L}_{\mathrm{str}}
\right\rangle
\right).
\end{aligned}
\]
Thus, $\rho_+\geq0$ measures the positive semantic--structural
cross-gradient interaction rather than conventional gradient
conflict. Coupled prototypes force semantic and structural cues into one
shared space and can therefore increase this interaction, whereas \DCPAab{}
is designed to reduce it through independent alignment.

\noindent\textbf{Theorem 1 (One-round surrogate descent bound).}
Under the conditions above, for a fixed aggregation vector $w$
within round $r$, the surrogate trajectory satisfies
\begin{align}
	\mathbb{E}\!\left[\mathcal{F}_w^{r+1}\right]
	&\leq
	\mathbb{E}\!\left[\mathcal{F}_w^{r}\right]
	-
	\alpha(\eta)
	\sum_{\ell=0}^{T-1}
	\mathbb{E}\!\left[G_{r,\ell}\right]
	\nonumber\\
	&\quad+
	\frac{L_{\mathrm{sm}}\eta^2T}{2}
	\bar{\sigma}_{w,\rho_+}^2
	+
	T\eta C_{\mathrm{pa}},
	\label{eq:thm1}
\end{align}
where
\[
\mathcal{F}_w^{r}=\mathcal{F}_w(\Theta_{r,0}),\qquad
\mathcal{F}_w^{r+1}=\mathcal{F}_w(\Theta_{r,T}),
\]
\[
G_{r,\ell}
=
\|\nabla \mathcal{F}_w(\Theta_{r,\ell})\|^2,
\qquad
\alpha(\eta)
=
\eta-\frac{L_{\mathrm{sm}}\eta^2}{2}.
\]
The result follows from the smooth non-convex descent lemma,
conditional unbiasedness, the effective second-moment bound,
and the assumed prototype-anchor perturbation bound. Detailed
derivations are provided in Appendix~C.

\noindent\textbf{Theorem 2 (Fixed-objective stationarity bound).}
Assume that the aggregation vector $w$ and hence the objective
$\mathcal{F}_w$ remain fixed over the $R$ analyzed rounds. Let
\[
\Delta
=
\mathbb{E}[\mathcal{F}_w^0]-\mathcal{F}_w^\star,
\]
where $\mathcal{F}_w^\star$ is a finite lower bound of $\mathcal{F}_w$. Then, for $\alpha(\eta)>0$,
\begin{align}
	\frac{1}{RT}
	\sum_{r=0}^{R-1}
	\sum_{\ell=0}^{T-1}
	\mathbb{E}[G_{r,\ell}]
	&\leq
	\frac{\Delta}{RT\alpha(\eta)}
	+
	\frac{L_{\mathrm{sm}}\eta^2}
	{2\alpha(\eta)}
		\bar{\sigma}_{w,\rho_+}^2
	\nonumber\\
	&\quad+
	\frac{\eta C_{\mathrm{pa}}}{\alpha(\eta)}.
\end{align}
Consequently, for a target $\epsilon>0$, the averaged squared
gradient norm is at most $\epsilon$ whenever
\begin{equation}
	R
	\geq
	\frac{2\Delta}
	{
		T\epsilon(2\eta-L_{\mathrm{sm}}\eta^2)
		-
		T L_{\mathrm{sm}}\eta^2\bar{\sigma}_{w,\rho_+}^2
		-
		2T\eta C_{\mathrm{pa}}
	},
	\label{eq:thm2}
\end{equation}
provided that the denominator is positive. A sufficient
condition is
\begin{equation}
	0<\eta<
	\frac{2(\epsilon-C_{\mathrm{pa}})}
	{L_{\mathrm{sm}}
		(\epsilon+\bar{\sigma}_{w,\rho_+}^2)},
	\qquad
	C_{\mathrm{pa}}<\epsilon.
\end{equation}

For a fixed nonzero learning rate, the last two terms in the
stationarity bound do not vanish as $R\rightarrow\infty$.
Therefore, the result guarantees convergence to a stationary
neighborhood of the surrogate objective rather than exact
stationarity, unless the learning rate and perturbation terms
vanish appropriately. For the round-varying \SSPAab{} weights,
Appendix~E of the supplementary material gives the same bound with an
extra weight-drift term $\Xi_R$.

\noindent\textbf{Implication for \oursplus{}.}
For comparison, we write the coupled \CDPAab{} loss in the same form,
$\lambda_{\text{DCPA}}(\mathcal{L}_{\text{sem}}+\lambda_{\text{str}}\mathcal{L}_{\text{str}})$,
with one shared prototype per class.
Let $\bar{\sigma}_{\text{conf}}^2=\sigma_a^2+
2\lambda_{\text{DCPA}}^2\lambda_{\text{str}}\rho_+^{\text{CDPA}}$ denote the effective second-moment term of the
conference FedBCS bound, where $a_m=N_m/\sum_jN_j$ are FedAvg weights. Let
$\bar{\sigma}_{+}^2=\sigma_{w^{\text{S2PA}}}^2+
2\lambda_{\text{DCPA}}^2\lambda_{\text{str}}\rho_+^{\text{DCPA}}$ denote the corresponding term for
\oursplus{}. If the prototype-guided weights reduce the aggregation-dependent term
($\sigma_{w^{\text{S2PA}}}^2\le\sigma_a^2$) and decoupled prototypes reduce
the positive semantic--structural cross-gradient interaction
($\rho_+^{\text{DCPA}}\le\rho_+^{\text{CDPA}}$), then
$\bar{\sigma}_{+}^2\le\bar{\sigma}_{\text{conf}}^2$. Consequently, the second-moment
term in \cref{eq:thm1} is reduced by
$\frac{L_{\mathrm{sm}}\eta^2T}{2}
(\bar{\sigma}_{\text{conf}}^2-\bar{\sigma}_{+}^2)$, and the denominator in
\cref{eq:thm2} increases by
$TL_{\mathrm{sm}}\eta^2(\bar{\sigma}_{\text{conf}}^2-\bar{\sigma}_{+}^2)$, yielding a
tighter convergence guarantee for \oursplus{} under these sufficient
conditions.

\section{Experiments}\label{sec:experiments}

In this section, we evaluate FedBCS+ on four 2D benchmarks and the 3D
FeTS~2022 benchmark. After describing the setup (Sec.~\ref{sec:setup}), we
compare FedBCS+ with state-of-the-art methods (Sec.~\ref{sec:comparison}),
analyze the contribution of each component (Sec.~\ref{sec:ablation}), and
examine its training dynamics, style separation, prototype entanglement,
overhead, and theoretical implications (Sec.~\ref{sec:further}).

\subsection{Experimental Setup}
\label{sec:setup}

\begin{figure*}[t]
\centering
\includegraphics[width=\linewidth]{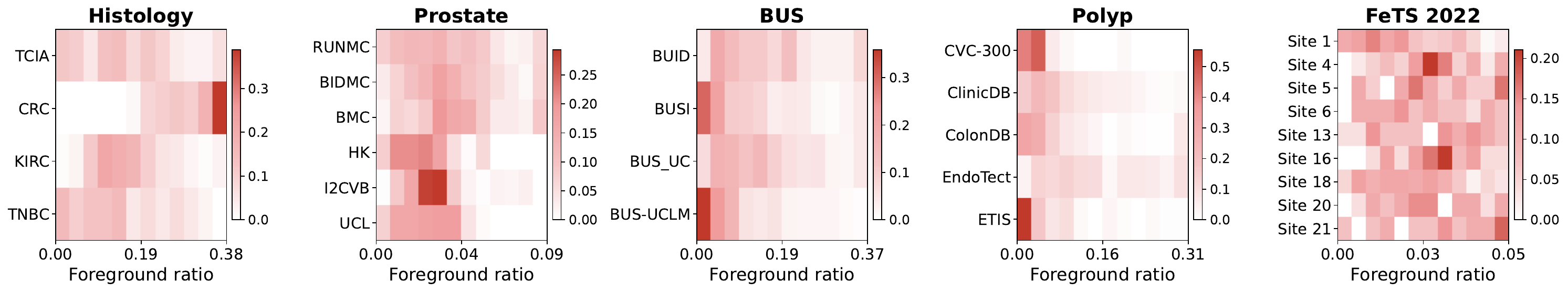}
\caption{\textbf{Foreground ratio distributions across clients.}
The four 2D benchmarks and the 3D FeTS~2022 benchmark exhibit substantial
variation in target size across clients. Each row represents a client,
and each column represents a foreground ratio bin. The distributions are
normalized separately for each client.}
\label{fig:client_het}
\end{figure*}

\begin{figure*}[t]
\centering
\includegraphics[width=\linewidth]{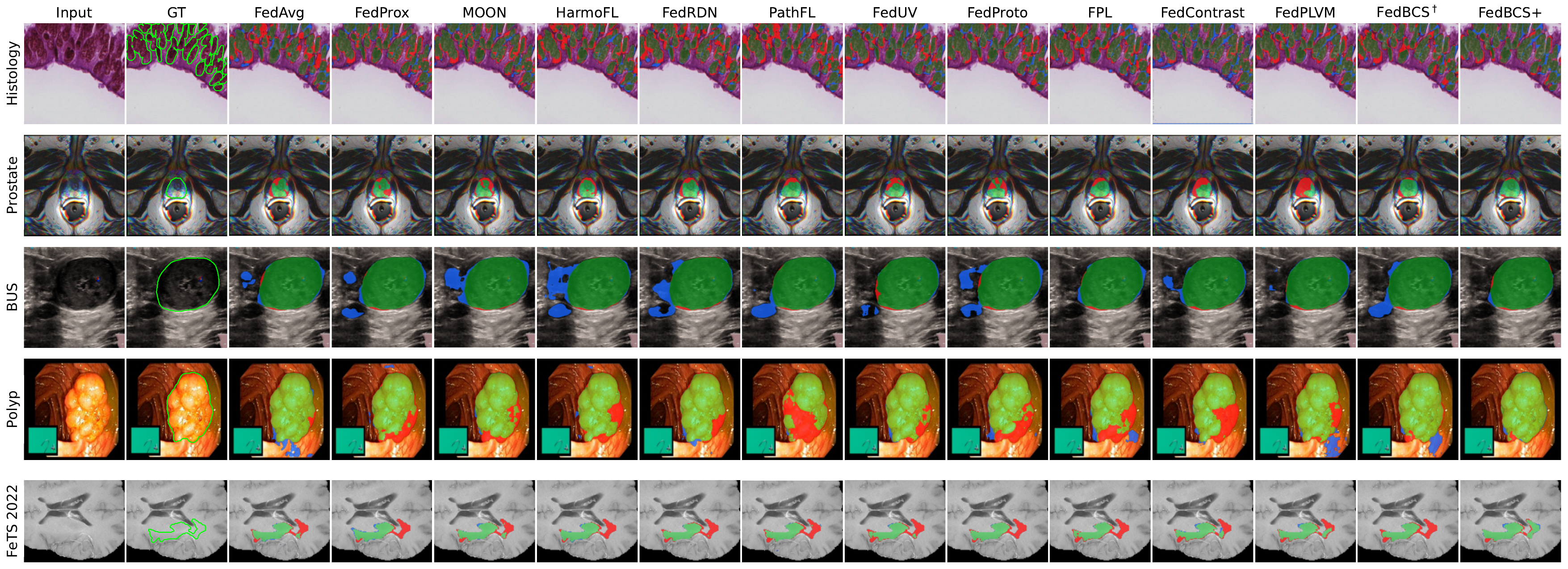}
\caption{\textbf{Qualitative segmentation results across the five benchmarks.}
Each row corresponds to a benchmark. The \emph{Input} column illustrates
appearance variations across imaging domains, and the \emph{GT} column
shows the ground-truth contours in \textcolor{green}{green}.
In the segmentation results, \textcolor{green}{green} indicates true positives,
\textcolor{red}{red} indicates false negatives, and \textcolor{blue}{blue}
indicates false positives. For FeTS~2022, the whole-tumor region is shown
in an axial slice, with the same slice and crop used for all methods.
FedBCS+ delineates the targets accurately with fewer missed regions
and false positives.}
\label{fig:example_cases}
\end{figure*}

\noindent\textbf{Datasets.}
We evaluate four heterogeneous 2D federated segmentation tasks, denoted
Histology, Prostate, BUS, and Polyp.
Histology follows the four-client nuclei segmentation setting of our
conference version~\cite{FedBCS_AAAI26}, comprising
TCIA~\cite{TCIA_Hou2020}, CRC, KIRC~\cite{KIRC_Irshad2014},
and TNBC~\cite{TNBC_Naylor2018}.
Prostate comprises six clients for T2-weighted MRI segmentation, drawn from
NCI-ISBI~2013, I2CVB, and PROMISE12~\cite{Nicholas2015,Lemaitre2015,Litjens2014}.
BUS comprises four clients for breast lesion segmentation in ultrasound:
the Breast Ultrasound Image Database from QAMEBI (BUID), BUSI, BUS\_UC,
and BUS-UCLM~\cite{BUID,BUSI_AlDhabyani2020,BUS_UC,BUS_UCLM}.
Polyp comprises five clients: the CVC-EndoSceneStill subset commonly referred
to as CVC-300~\cite{CVC300_Vazquez2017},
CVC-ClinicDB (ClinicDB)~\cite{ClinicalDB_Bernal2015},
CVC-ColonDB (ColonDB)~\cite{CVCColonDB_Bernal2012},
the EndoTect~2020 segmentation test set (EndoTect)~\cite{EndoTect_Hasan2021},
and ETIS-LaribPolypDB (ETIS)~\cite{ETIS_Silva2014}.
We additionally evaluate 3D brain tumor segmentation using nine institutional
partitions from the Federated Tumor Segmentation (FeTS) 2022 data
release~\cite{FeTS_dataset,FeTS2022_data}.
Table~\ref{tab:datasets} lists the client IDs, imaging modalities, and segmentation tasks.
Fig.~\ref{fig:client_het} quantifies heterogeneity in target size using the foreground ratio, and
Fig.~\ref{fig:example_cases} shows an example input from each benchmark.
Client datasets range from 50 to 811 images (22 to 511 volumes for
FeTS~2022) and differ substantially in their foreground-ratio distributions.

\begin{table*}[t]\footnotesize
\caption{\textbf{Datasets and client configurations.}
$M$ denotes the number of federated clients used for each benchmark.}
\label{tab:datasets}
\vspace{-6pt}
\centering
{\setlength\tabcolsep{3.2pt}
\renewcommand\arraystretch{1.06}
\begin{tabular}{
  >{\centering\arraybackslash}m{2.35cm}|
  >{\centering\arraybackslash}m{0.35cm}|
  >{\centering\arraybackslash}m{1.85cm}|
  >{\centering\arraybackslash}m{2.15cm}|
  >{\centering\arraybackslash}m{257.0258pt}|
  >{\centering\arraybackslash}m{1.05cm}}
\hline\thickhline
\rowcolor{mydarkyellow}
\textbf{Benchmark} & $M$ & \textbf{Modality} & \textbf{Task}
  & \textbf{Client IDs} & \textbf{Example} \\
\hline
\rowcolor{gray!10}
\multicolumn{6}{l}{\textit{2D benchmarks}} \\
\hline
Histology
  & 4 & Histology & Nuclei
  & TCIA, CRC, KIRC, TNBC~\cite{TCIA_Hou2020,KIRC_Irshad2014,TNBC_Naylor2018}
  & \adjustbox{valign=c}{\includegraphics[height=0.76cm]{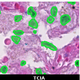}} \\
Prostate
  & 6 & T2w MRI & Prostate gland
  & RUNMC, BIDMC, BMC, HK, I2CVB, UCL~\cite{Nicholas2015,Lemaitre2015,Litjens2014}
  & \adjustbox{valign=c}{\includegraphics[height=0.76cm]{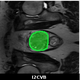}} \\
BUS
  & 4 & Ultrasound & Breast lesion
  & BUID, BUSI, BUS\_UC, BUS-UCLM~\cite{BUID,BUSI_AlDhabyani2020,BUS_UC,BUS_UCLM}
  & \adjustbox{valign=c}{\includegraphics[height=0.76cm]{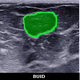}} \\
Polyp
  & 5 & Colonoscopy & Polyp
  & CVC-300, ClinicDB, ColonDB, EndoTect, ETIS~\cite{CVC300_Vazquez2017,ClinicalDB_Bernal2015,CVCColonDB_Bernal2012,EndoTect_Hasan2021,ETIS_Silva2014}
  & \adjustbox{valign=c}{\includegraphics[height=0.76cm]{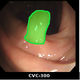}} \\
\hline\thickhline
\rowcolor{gray!10}
\multicolumn{6}{l}{\textit{3D benchmark}} \\
\hline
FeTS~2022
  & 9 & MRI & Tumor regions
  & Sites 1,4,5,6,13,16,18,20,21~\cite{FeTS2022_data}
  & \adjustbox{valign=c}{\includegraphics[height=0.76cm]{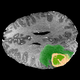}} \\
\hline\thickhline
\end{tabular}}
\end{table*}

\noindent\textbf{Evaluation Metrics and Protocol.}
We use the Dice score, expressed as a percentage, to measure region overlap,
and the standard deviation $\sigma_{\text{Dice}}$ of the per-client mean Dice
scores to measure cross-client consistency. The main experiments on the four
2D benchmarks use five-fold cross-validation: each client's dataset is randomly
partitioned into five folds with a fixed seed, the same fold index is held out
for testing on all clients, and we report the mean and standard deviation over
the five runs. The 3D FeTS~2022 and leave-one-site-out (LOSO) experiments use
the first split. In LOSO, which evaluates generalization to unseen domains, one
client is excluded from federated training and evaluated without fine-tuning
at the round with the highest mean Dice over the remaining clients.

\noindent\textbf{Baselines.}
We compare FedBCS+ with representative federated learning methods
in two categories, as well as with our conference version. Table~\ref{tab:method_analysis} compares FedAvg and the
prototype-based methods in model size, computational cost, and prototype
communication cost on Histology.

\noindent\textit{General FL Methods.}
FedAvg~\cite{FedAvg_AISTATS17} averages locally trained parameters to form the global model.
FedProx~\cite{FedProx_MLSys2020} adds a proximal regularization term
to constrain local updates.
MOON~\cite{MOON_CVPR21} applies model-level contrastive learning
to align local and global representations.
HarmoFL~\cite{jiang2022harmofl} combines amplitude normalization
with weight perturbation to reduce local and global drifts.
FedRDN~\cite{yan2025simple} incorporates global feature statistics into local samples through random data normalization to address feature-distribution skew.
PathFL~\cite{zhang2025pathfl} performs cross-site alignment at the image, feature, and aggregation levels for heterogeneous segmentation.
FedUV~\cite{FedUV_CVPR24} regularizes training by promoting representation uniformity and variance in classifier outputs.

\noindent\textit{Prototype-based FL Methods.}
FedProto~\cite{FedProto_AAAI22} communicates class-level prototypes across clients.
FPL~\cite{FPL_CVPR23} employs cluster-based prototypes with contrastive learning.
FedContrast~\cite{FedContrast-GPA_MICCAI23}, \ie, FedContrast-GPA without its process-aware aggregation, uses prototype-based contrastive learning for local discrimination and cross-client representation alignment.
FedPLVM~\cite{FedPLVM_NeurIPS24} clusters features to construct multiple prototypes per class.

\noindent\textit{Conference Version.}
FedBCS~\cite{FedBCS_AAAI26} is our conference version, which combines FSR and CDPA.

\begin{table}[t]\small
\caption{\textbf{Client-side cost and segmentation accuracy on Histology.}
$P$ denotes the model parameter count, and $N_{\text{proto}}$ is the number
of prototypes uploaded per client per round. FLOPs are computed for a
$256 \times 256$ input, including the FFT operations. The $\Delta$ columns
report percentage increases relative to UNet. FSR adds 902 parameters and
increases FLOPs by 0.98\%. A dagger denotes our conference version.}
\label{tab:method_analysis}
\vspace{-10pt}
\centering
{\setlength\tabcolsep{4.5pt}
\renewcommand\arraystretch{1.2}%
\resizebox{\columnwidth}{!}{%
\begin{tabular}{lIcc|cc|c|c}
\hline\thickhline
\rowcolor{mydarkyellow}
\textbf{Method} & \textbf{$P$\,(M)} & \textbf{$\Delta$\,(\%)}
  & \textbf{FLOPs\,(G)} & \textbf{$\Delta$\,(\%)}
  & \textbf{$N_{\text{proto}}$} & \textbf{Dice\,(\%)} \\
\hline\hline
FedAvg       & 8.267 & --    & 24.147 & --    & $0$ & 71.14 \\
\rowcolor{mylightyellow}
FedProto        & 8.267 & --    & 24.147 & --    & $2$ & 70.56 \\
FPL             & 8.267 & --    & 24.147 & --    & $2$ & 72.25 \\
\rowcolor{mylightyellow}
FedContrast     & 8.268 & +0.01 & 24.383 & +0.98 & $9$ & 56.90 \\
FedPLVM      & 8.267 & --    & 24.147 & --    & $\approx 212$ & 70.64 \\
\rowcolor{mylightyellow}
FedBCS$^\dagger$& 8.268 & +0.01 & 24.383 & +0.98 & $4$ & 73.64 \\
\rowcolor{mydarkyellow}
\textbf{FedBCS+}              & 8.268 & +0.01 & 24.383 & +0.98 & $\leq 6$ & \textbf{75.47} \\
\hline\thickhline
\end{tabular}}}
\end{table}

\noindent\textbf{Model Architecture.}
For the four 2D benchmarks, we use UNet~\cite{ronneberger2015u} as the shared
segmentation backbone. The encoder contains four downsampling blocks, and
the decoder mirrors this structure with skip connections. For 3D FeTS~2022,
we use DynUNet~\cite{isensee2021nnu} with $128 \times 128 \times 128$ input patches to process volumetric MRI.
Within each benchmark, all clients share the same architecture and
synchronize their model parameters through federated aggregation.

\noindent\textbf{Federated Training Configuration.}
Following our conference version~\cite{FedBCS_AAAI26}, each client performs
one local epoch per communication round.
For Histology, we use SGD with a learning rate of $1\times10^{-2}$.
For Prostate, BUS, and Polyp, we use Adam with a learning rate of
$1\times10^{-4}$, and for FeTS~2022, Adam with a learning rate of
$2\times10^{-4}$.
We use a weight decay of $1\times10^{-4}$ and a training batch size of
six throughout. The learning rate is kept constant during training,
except for FedProx, whose implementation applies polynomial decay.
We fix the random seed across experiments for reproducibility.

\noindent\textbf{Module Hyperparameters.}
\textit{FSR.}
The logits $\mathbf{h}$ of \cref{eq:learnable_weight} are initialized to zero,
so that both mixing weights start at $0.5$, and the temperature is $T_s=0.1$.
We apply FSR to the three deeper encoder stages and retain the original
features at the shallowest stage.
\textit{DCPA.}
We select $K=4$ feature levels in each pathway. The erosion radius is
$\delta=2$ pixels at the label resolution, and the resulting masks are resized
to each feature map by nearest-neighbor interpolation. For FeTS~2022, the nested
tumor core (TC), whole tumor (WT), and enhancing tumor (ET) labels are first
converted into three
mutually exclusive foreground classes (edema, necrotic core, and enhancing
tumor). The loss weights are $\lambda_{\mathrm{DCPA}}=1.0$ and
$\lambda_{\text{str}}=0.05$, and the contrastive temperature $\tau$ is selected
per benchmark from $\{0.005, 0.01, 0.05, 0.1, 0.2\}$. Server-side prototype
clustering uses FINCH~\cite{sarfraz2019efficient}.
\textit{S2PA.}
The momentum coefficient $\beta_{\mathrm{ema}}$ is tuned per benchmark.

\noindent\textbf{Hardware.}
All experiments are conducted on NVIDIA A100 GPUs, each with 80\,GB of memory.

\begin{table*}[!t]
\caption{\textbf{Comparison with state-of-the-art federated learning methods on
four 2D medical segmentation benchmarks.}
Dice scores are expressed as percentages and reported as means and standard
deviations over five folds. The Avg. columns report results averaged across
clients.
The best results are shown in \textbf{bold} and the second best are
\underline{underlined}. A dagger denotes our conference
version~\cite{FedBCS_AAAI26}.}
\label{tab:compare_sota}
\vspace{-10pt}
\centering
\scriptsize

{\setlength\tabcolsep{4pt}
\renewcommand\arraystretch{1.2}%
\resizebox{\linewidth}{!}{%
\begin{tabular}{
>{\columncolor{mylightgray}}r
||
>{\columncolor{lightpink}}c
>{\columncolor{lightpink}}c
>{\columncolor{lightpink}}c
>{\columncolor{lightpink}}c
>{\columncolor{lightpink}}c
I
>{\columncolor{lightblue}}c
>{\columncolor{lightblue}}c
>{\columncolor{lightblue}}c
>{\columncolor{lightblue}}c
>{\columncolor{lightblue}}c
>{\columncolor{lightblue}}c
>{\columncolor{lightblue}}c
}
\hline\thickhline
\rowcolor{gray!20}
& \multicolumn{5}{cI}{\textbf{Histology} (4 clients)}
& \multicolumn{7}{c}{\textbf{Prostate} (6 clients)}
\\
\cline{2-13}
\rowcolor{gray!20}
\textbf{Method}
& TCIA & CRC & KIRC & TNBC & Avg.
& RUNMC & BIDMC & BMC & HK & I2CVB & UCL & Avg.
\\
\hline\hline

\multicolumn{13}{l}{\textbf{\textit{General FL Methods}}}
\\
FedAvg
& 71.78\spm{3.20} & 75.38\spm{1.77} & 73.19\spm{0.55} & 64.22\spm{4.17} & 71.14\spm{1.10} & 82.29\spm{2.10} & 71.79\spm{3.84} & 82.50\spm{2.26} & 81.74\spm{3.06} & 82.99\spm{3.25} & 73.55\spm{2.70} & 79.14\spm{1.44}
\\
FedProx
& 70.45\spm{3.48} & 75.08\spm{1.09} & 72.33\spm{1.89} & 63.75\spm{5.26} & 70.40\spm{0.43} & 80.92\spm{2.01} & 71.37\spm{5.36} & 80.52\spm{2.45} & 81.31\spm{2.99} & 81.01\spm{3.60} & 71.55\spm{4.31} & 77.78\spm{1.71}
\\
MOON
& 70.87\spm{3.35} & 74.66\spm{0.98} & 72.21\spm{1.85} & 64.26\spm{4.24} & 70.50\spm{0.59} & 82.83\spm{2.17} & 72.09\spm{6.04} & 81.83\spm{1.75} & 81.78\spm{4.18} & 82.78\spm{3.53} & 74.74\spm{5.52} & 79.34\spm{1.29}
\\
HarmoFL
& 74.40\spm{2.62} & 70.94\spm{2.54} & 72.09\spm{0.76} & 62.57\spm{3.03} & 70.00\spm{1.65} & 84.45\spm{2.13} & \underline{77.56}\spm{5.77} & 83.74\spm{2.17} & 82.60\spm{3.48} & 84.47\spm{2.93} & 76.75\spm{2.64} & 81.59\spm{1.60}
\\
FedRDN
& 71.58\spm{3.21} & 76.56\spm{1.97} & 73.18\spm{0.56} & 64.10\spm{3.52} & 71.36\spm{1.14} & 81.46\spm{1.52} & 71.55\spm{4.92} & 82.24\spm{1.08} & 81.38\spm{3.36} & 82.68\spm{4.69} & 71.74\spm{4.16} & 78.51\spm{1.64}
\\
PathFL
& 73.64\spm{3.99} & 75.67\spm{2.49} & 72.91\spm{1.01} & \underline{69.99}\spm{1.62} & 73.05\spm{0.99} & 82.49\spm{4.07} & \textbf{77.89}\spm{4.82} & \textbf{85.01}\spm{2.05} & 84.50\spm{3.16} & 83.99\spm{3.06} & \underline{77.77}\spm{4.29} & \underline{81.94}\spm{1.16}
\\
FedUV
& \textbf{75.43}\spm{3.83} & \underline{78.76}\spm{2.03} & \underline{74.38}\spm{0.71} & 66.21\spm{1.60} & \underline{73.70}\spm{1.09} & 81.62\spm{2.40} & 71.90\spm{4.22} & 82.15\spm{2.28} & 80.18\spm{2.68} & 80.96\spm{4.04} & 70.93\spm{3.85} & 77.96\spm{1.31}
\\
\hdashline
\multicolumn{13}{l}{\textbf{\textit{Prototype-based FL Methods}}}
\\
FedProto
& 70.64\spm{2.53} & 74.77\spm{2.71} & 72.86\spm{1.01} & 63.96\spm{4.23} & 70.56\spm{1.38} & 80.63\spm{2.23} & 69.50\spm{7.33} & 81.54\spm{2.64} & 80.27\spm{2.42} & 82.55\spm{2.15} & 70.91\spm{4.51} & 77.57\spm{2.25}
\\
FPL
& 71.93\spm{3.07} & 77.29\spm{2.05} & 74.05\spm{0.41} & 65.73\spm{4.39} & 72.25\spm{0.93} & 83.91\spm{1.77} & 75.33\spm{5.87} & 83.76\spm{1.56} & 81.09\spm{4.75} & 83.80\spm{2.98} & 73.89\spm{3.71} & 80.30\spm{1.31}
\\
FedContrast
& 58.26\spm{2.55} & 67.27\spm{3.79} & 56.38\spm{0.70} & 45.68\spm{5.73} & 56.90\spm{1.70} & 73.65\spm{4.53} & 65.65\spm{6.62} & 75.03\spm{2.14} & 79.40\spm{3.34} & 76.83\spm{2.61} & 63.29\spm{8.32} & 72.31\spm{1.13}
\\
FedPLVM
& 73.41\spm{1.62} & 71.20\spm{1.88} & 73.44\spm{0.28} & 64.50\spm{3.12} & 70.64\spm{0.99} & 79.16\spm{4.57} & 68.72\spm{4.18} & 81.00\spm{3.29} & 79.96\spm{4.89} & 81.74\spm{3.91} & 71.00\spm{5.77} & 76.93\spm{1.63}
\\
\hdashline
\multicolumn{13}{l}{\textbf{\textit{Ours}}}
\\
FedBCS$^\dagger$
& 74.16\spm{4.79} & 77.50\spm{1.41} & \textbf{74.68}\spm{0.48} & 68.21\spm{1.64} & 73.64\spm{1.30} & \underline{84.83}\spm{0.88} & 74.12\spm{3.88} & \underline{84.61}\spm{1.61} & \textbf{85.40}\spm{1.36} & \textbf{86.37}\spm{1.29} & 75.17\spm{1.50} & 81.75\spm{1.10}
\\
\rowcolor{oursrow}
\textbf{FedBCS+}
& \underline{75.28}\spm{3.99} & \textbf{81.14}\spm{1.32} & 73.80\spm{0.70} & \textbf{71.68}\spm{1.60} & \textbf{75.47}\spm{0.86} & \textbf{85.14}\spm{1.56} & 75.66\spm{7.01} & 83.61\spm{2.11} & \underline{84.85}\spm{2.73} & \underline{85.34}\spm{2.94} & \textbf{80.08}\spm{1.84} & \textbf{82.45}\spm{1.16}
\\
\hline\thickhline
\end{tabular}}}

\vspace{3pt}

{\setlength\tabcolsep{4pt}
\renewcommand\arraystretch{1.2}%
\resizebox{\linewidth}{!}{%
\begin{tabular}{
>{\columncolor{mylightgray}}r
||
>{\columncolor{lightgreen}}c
>{\columncolor{lightgreen}}c
>{\columncolor{lightgreen}}c
>{\columncolor{lightgreen}}c
>{\columncolor{lightgreen}}c
I
>{\columncolor{lightyellow}}c
>{\columncolor{lightyellow}}c
>{\columncolor{lightyellow}}c
>{\columncolor{lightyellow}}c
>{\columncolor{lightyellow}}c
>{\columncolor{lightyellow}}c
}
\hline\thickhline
\rowcolor{gray!20}
& \multicolumn{5}{cI}{\textbf{BUS} (4 clients)}
& \multicolumn{6}{c}{\textbf{Polyp} (5 clients)}
\\
\cline{2-12}
\rowcolor{gray!20}
\textbf{Method}
& BUID & BUSI & BUS\_UC & BUS-UCLM & Avg.
& CVC-300 & ClinicDB & ColonDB & EndoTect & ETIS & Avg.
\\
\hline\hline

\multicolumn{12}{l}{\textbf{\textit{General FL Methods}}}
\\
FedAvg
& 80.98\spm{2.47} & 71.18\spm{3.14} & 88.75\spm{0.77} & 67.33\spm{1.82} & 77.06\spm{1.09} & 94.93\spm{1.52} & 78.93\spm{3.48} & 80.18\spm{2.72} & 68.44\spm{2.27} & 65.12\spm{3.56} & 77.52\spm{1.19}
\\
FedProx
& 81.76\spm{2.30} & 70.34\spm{2.75} & 88.28\spm{1.04} & 67.12\spm{2.91} & 76.87\spm{1.10} & 93.34\spm{2.48} & 78.75\spm{3.06} & 77.79\spm{3.39} & 67.44\spm{2.35} & 61.95\spm{3.50} & 75.85\spm{1.04}
\\
MOON
& 79.61\spm{1.96} & 69.55\spm{3.51} & 87.75\spm{1.27} & 65.01\spm{3.99} & 75.48\spm{0.94} & 93.81\spm{1.74} & 79.76\spm{3.01} & 77.65\spm{3.42} & 68.97\spm{2.64} & 66.68\spm{2.36} & 77.37\spm{0.59}
\\
HarmoFL
& \underline{81.81}\spm{2.16} & \underline{71.76}\spm{2.80} & \textbf{89.66}\spm{0.42} & 69.05\spm{2.30} & \underline{78.07}\spm{0.93} & \textbf{95.90}\spm{1.36} & \textbf{82.20}\spm{3.73} & \textbf{82.59}\spm{3.32} & \underline{73.49}\spm{2.38} & 67.21\spm{5.16} & \underline{80.28}\spm{1.51}
\\
FedRDN
& 80.76\spm{1.79} & 70.67\spm{2.70} & 87.74\spm{0.85} & 66.87\spm{3.47} & 76.51\spm{1.37} & 95.02\spm{1.29} & 79.21\spm{3.32} & 79.59\spm{3.57} & 70.19\spm{1.71} & 62.78\spm{4.72} & 77.36\spm{1.41}
\\
PathFL
& 80.75\spm{2.05} & 69.33\spm{3.68} & 88.13\spm{1.22} & 69.85\spm{1.39} & 77.02\spm{0.80} & 95.43\spm{1.27} & 76.95\spm{2.96} & \underline{81.80}\spm{4.01} & \textbf{74.34}\spm{2.66} & \underline{69.99}\spm{4.61} & 79.70\spm{1.44}
\\
FedUV
& 81.62\spm{2.58} & 69.75\spm{2.72} & 88.62\spm{0.83} & 67.03\spm{3.27} & 76.76\spm{0.96} & \underline{95.59}\spm{1.02} & 80.48\spm{4.00} & 81.78\spm{3.79} & 71.44\spm{0.23} & 62.82\spm{4.61} & 78.42\spm{1.09}
\\
\hdashline
\multicolumn{12}{l}{\textbf{\textit{Prototype-based FL Methods}}}
\\
FedProto
& 80.54\spm{2.79} & 70.29\spm{3.25} & 88.06\spm{1.07} & 68.74\spm{2.02} & 76.91\spm{1.70} & 91.88\spm{1.48} & 77.23\spm{3.51} & 75.38\spm{4.90} & 66.95\spm{1.27} & 58.75\spm{4.42} & 74.04\spm{1.65}
\\
FPL
& 81.34\spm{2.80} & 70.71\spm{3.18} & 89.15\spm{0.55} & 68.89\spm{2.84} & 77.52\spm{1.18} & 94.46\spm{1.25} & 80.47\spm{3.29} & 80.17\spm{3.49} & 71.16\spm{2.24} & 66.29\spm{3.19} & 78.51\spm{0.86}
\\
FedContrast
& 79.23\spm{1.36} & 64.14\spm{3.20} & 87.10\spm{0.66} & 63.39\spm{4.46} & 73.46\spm{1.02} & 88.75\spm{2.08} & 66.48\spm{2.33} & 69.85\spm{4.91} & 55.74\spm{2.13} & 56.39\spm{2.04} & 67.44\spm{0.86}
\\
FedPLVM
& 80.94\spm{1.99} & 70.39\spm{2.54} & 88.72\spm{0.85} & 66.86\spm{2.21} & 76.73\spm{0.99} & 92.64\spm{0.90} & 76.05\spm{3.05} & 75.58\spm{4.56} & 65.37\spm{4.01} & 56.31\spm{2.71} & 73.19\spm{1.06}
\\
\hdashline
\multicolumn{12}{l}{\textbf{\textit{Ours}}}
\\
FedBCS$^\dagger$
& 81.20\spm{2.37} & 71.28\spm{3.09} & \underline{89.26}\spm{0.66} & \underline{70.48}\spm{1.74} & 78.06\spm{1.33} & 95.29\spm{0.84} & \underline{82.11}\spm{2.33} & 81.67\spm{3.95} & 70.81\spm{2.51} & 67.61\spm{4.36} & 79.50\spm{0.93}
\\
\rowcolor{oursrow}
\textbf{FedBCS+}
& \textbf{82.12}\spm{2.28} & \textbf{72.65}\spm{3.09} & 88.60\spm{0.90} & \textbf{73.53}\spm{2.01} & \textbf{79.22}\spm{1.13} & 94.69\spm{1.06} & 80.60\spm{1.41} & 81.39\spm{3.16} & 73.03\spm{1.88} & \textbf{74.19}\spm{3.22} & \textbf{80.78}\spm{1.09}
\\
\hline\thickhline
\end{tabular}}}
\end{table*}

\begin{table}[t]\small
\caption{\textbf{Brain tumor segmentation on FeTS~2022.}
The experiment uses nine clients and the first cross-validation split.
TC, WT, and ET denote tumor core, whole tumor, and enhancing tumor,
respectively. Dice scores are expressed as percentages, and the final
column reports the mean across the three tumor regions. The best results
are shown in \textbf{bold} and the second best are \underline{underlined}.
A dagger denotes our conference version.}
\label{tab:compare_brats}
\vspace{-10pt}
\centering
{\setlength\tabcolsep{4pt}
\renewcommand\arraystretch{1.1}%
\begin{tabular}{rIc|c|cIc}
\hline\thickhline
\rowcolor{mydarkred}
\textbf{Method} & \textbf{TC} & \textbf{WT} & \textbf{ET}
  & \textbf{Avg.\ Dice} \\
\hline\hline
\multicolumn{5}{l}{\textbf{\textit{General FL Methods}}} \\
\hline
FedAvg
  & 80.10 & 91.24 & 81.47 & 84.27
\\
\rowcolor{mylightred}
FedProx
  & 80.94 & 90.78 & 81.86 & 84.53
\\
MOON
  & 80.49 & 91.29 & \underline{82.14} & 84.64
\\
\rowcolor{mylightred}
HarmoFL
  & 80.31 & \textbf{91.48} & 81.28 & 84.36
\\
FedRDN
  & 80.42 & 91.03 & 81.79 & 84.42
\\
\rowcolor{mylightred}
PathFL
  & 81.19 & 90.06 & 81.43 & 84.23
\\
FedUV
  & \underline{81.21} & 90.97 & 81.12 & 84.43
\\
\hdashline
\multicolumn{5}{l}{\textbf{\textit{Prototype-based FL Methods}}} \\
\hline
FedProto
  & 80.81 & \textbf{91.48} & 81.76 & \underline{84.68}
\\
\rowcolor{mylightred}
FPL
  & 77.44 & 89.33 & 79.11 & 81.96
\\
FedContrast
  & 80.71 & 90.39 & 81.35 & 84.15
\\
\rowcolor{mylightred}
FedPLVM
  & 80.64 & 91.00 & 81.10 & 84.24
\\
\hdashline
\multicolumn{5}{l}{\textbf{\textit{Ours}}} \\
\hline
FedBCS$^\dagger$
  & 80.12 & 90.29 & \textbf{82.19} & 84.20
\\
\rowcolor{oursrow}
\textbf{FedBCS+}
  & \textbf{82.73} & \underline{91.45} & 81.71 & \textbf{85.30} \\
\hline\thickhline
\end{tabular}}
\end{table}

\subsection{Comparison with State-of-the-art Methods}
\label{sec:comparison}

Table~\ref{tab:compare_sota} compares performance on the four 2D benchmarks
under five-fold cross-validation. FedBCS+ achieves the highest mean Dice
scores of 75.47\%, 82.45\%, 79.22\%, and 80.78\% on Histology, Prostate,
BUS, and Polyp, respectively. In the additional 3D evaluation on
FeTS~2022, FedBCS+ also achieves the highest mean Dice score of 85.30\%,
as shown in Table~\ref{tab:compare_brats}. It outperforms the conference
version, FedBCS, on all five benchmarks.

\noindent\textbf{Segmentation Accuracy.}
As shown in Table~\ref{tab:compare_sota},
FedBCS+ improves over FedAvg by 4.33, 3.31, 2.16, and 3.26 Dice points on
the four main benchmarks, respectively. It also improves Dice by 1.03 points
on 3D FeTS~2022 (Table~\ref{tab:compare_brats}).
General FL methods, including FedAvg, MOON, and HarmoFL, lack explicit
cross-client prototype alignment. Prototype-based methods, including
FedProto, FPL, and FedPLVM, align client representations but rely on
single-layer features and conventional aggregation. FedBCS+ targets both
limitations by combining the multi-level contextual alignment of DCPA with
the prototype-guided weighting of S2PA.
The qualitative examples in Fig.~\ref{fig:example_cases} are consistent
with the quantitative results, with fewer missed regions and false positives for FedBCS+.

\noindent\textbf{Cross-client Consistency and Generalization.}
FedBCS+ also improves performance consistency across clients. Relative to
FedAvg, $\sigma_{\text{Dice}}$ decreases on all four 2D benchmarks: from 4.20
to 3.51 on Histology, 4.62 to 3.52 on Prostate, 8.39 to 6.56 on BUS, and
10.47 to 7.71 on Polyp. FedBCS+ attains the lowest $\sigma_{\text{Dice}}$
among all methods on BUS and Polyp and the third lowest on Histology and
Prostate. This agrees with the design of S2PA, which gives more weight to
clients that are under-represented in the global consensus.
Table~\ref{tab:loso} reports the LOSO results on Histology. FedBCS+ achieves
the highest average Dice score of 69.97\%, compared with 67.92\% for
FedBCS and at most 67.14\% for the other methods in Table~\ref{tab:loso}. These results indicate better
transfer to unseen imaging domains.

\noindent\textbf{Comparison with the Conference Version.}
FedBCS+ extends FedBCS in two respects.
\textbf{(1)~Decoupled prototype alignment.} DCPA replaces coupled prototype alignment with
separate semantic and structural branches, so that regional and contour cues
are aligned separately. In the standalone alignment comparison on
Histology (Sec.~\ref{sec:dcpa_fine}), this change increases Dice from
71.50\% to 72.23\%.
\textbf{(2)~Adaptive parameter aggregation.} Weighting clients by their
semantic--structural prototype divergence provides the largest standalone
gain of 2.67 Dice points in Table~\ref{tab:component_ablation}.
Together, these extensions yield gains over FedBCS ranging from 0.70 Dice
points on Prostate to 1.83 points on Histology, with 1.10 points on
FeTS~2022.

\begin{figure*}[t]
\centering
\includegraphics[width=\textwidth]{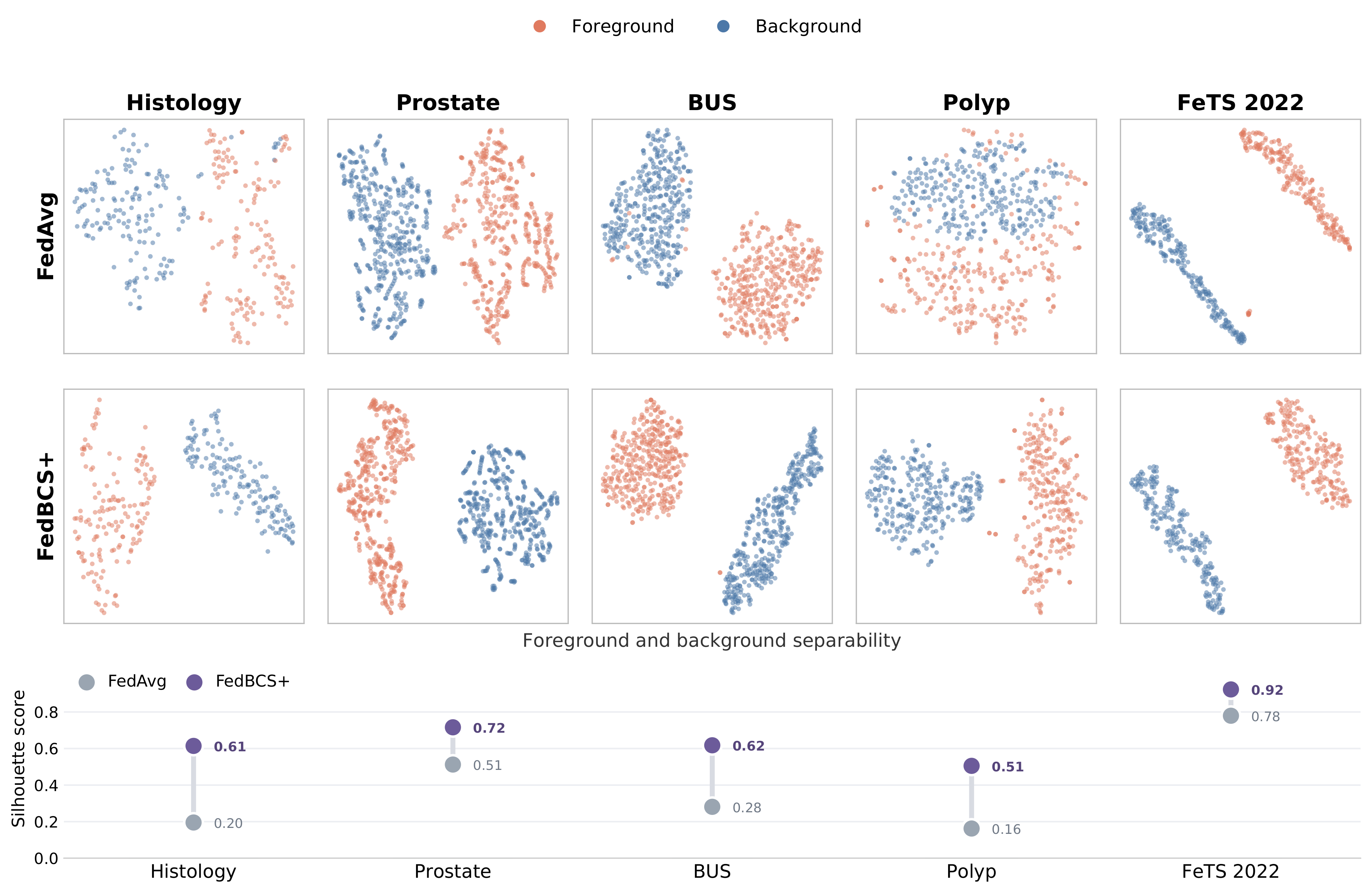}
\caption{\textbf{Foreground and background feature separability.}
The upper panels show t-SNE visualizations of encoder features learned by
FedAvg and FedBCS+ on the four 2D benchmarks and FeTS~2022. The lower panel
compares their silhouette scores. FedBCS+ achieves higher scores across all
five benchmarks, indicating clearer separation between foreground and
background features.}
\label{fig:tsne_features}
\end{figure*}

\noindent\textbf{Feature Separability.}
Fig.~\ref{fig:tsne_features} visualizes pixel-level foreground and background
encoder features using t-SNE~\cite{tsne_JMLR08} to examine the representations learned by FedBCS+.
Compared with FedAvg, FedBCS+ produces more separable features on every
benchmark. The silhouette score rises most on Histology (from 0.20 to 0.61)
and Polyp (from 0.16 to 0.51), and least on FeTS~2022 (from 0.78 to 0.92),
where the FedAvg features are already well separated.

\begin{table}[t]\small
\caption{\textbf{Leave-one-site-out generalization on Histology.}
Each site is excluded from federated training in turn and evaluated
without fine-tuning at the round with the highest mean Dice over the three
training sites. The experiment uses the first cross-validation split.
Dice scores are expressed as percentages,
and Avg. denotes the mean across held-out sites. The best results are
shown in \textbf{bold} and the second best are \underline{underlined}.
A dagger denotes our conference version.}
\label{tab:loso}
\vspace{-10pt}
\centering
{\setlength\tabcolsep{5pt}
\renewcommand\arraystretch{1.2}%
\resizebox{0.85\columnwidth}{!}{%
\begin{tabular}{rIccccIc}
\hline\thickhline
\rowcolor{mydarkblue}
\textbf{Method} & TCIA & CRC & KIRC & TNBC & Avg. \\
\hline\hline
\rowcolor{mylightblue}
FedAvg         & 60.45 & 72.83 & \underline{70.69} & 64.58 & 67.14 \\
FedProx        & 60.22 & 72.85 & 69.65 & 64.90 & 66.91 \\
\rowcolor{mylightblue}
MOON           & 60.38 & 72.89 & 70.68 & 64.59 & 67.14 \\
FPL            & 59.25 & \underline{73.55} & 69.22 & 66.30 & 67.08 \\
\rowcolor{mylightblue}
FedPLVM        & 68.89 & 59.65 & 70.64 & \underline{68.60} & 66.95 \\
FedBCS$^\dagger$ & \underline{70.63} & 59.54 & \textbf{71.58} & \textbf{69.93} & \underline{67.92} \\
\rowcolor{mydarkblue}
\textbf{FedBCS+}  & \textbf{71.23} & \textbf{74.01} & 70.64 & 63.99 & \textbf{69.97} \\
\hline\thickhline
\end{tabular}}}
\end{table}

\subsection{Ablation Studies}
\label{sec:ablation}

We conduct ablation studies on Histology to evaluate the contribution of
each component. The component ablation reports averages over five folds;
the fold settings for the remaining analyses are specified in their captions.

\subsubsection{Component Ablation}

Table~\ref{tab:component_ablation} evaluates the three components in
two settings: \emph{individual}, where each module is added separately to
FedAvg, and \emph{cumulative}, where the modules are progressively combined
to form FedBCS+.
\begin{table}[t]\small
\caption{\textbf{Component ablation on Histology.}
Dice scores are expressed as percentages and reported as means and standard
deviations over five folds. The upper block evaluates each module added
individually to FedAvg. The lower block progressively combines the modules
to form FedBCS+. The best results are shown in \textbf{bold} and the second
best are \underline{underlined}.}
\label{tab:component_ablation}
\vspace{-10pt}
\centering
\small
\setlength\tabcolsep{4pt}
\renewcommand\arraystretch{1.2}
\begin{tabular}{lIcccIc}
\hline\thickhline
\rowcolor{mydarkred}
\textbf{Setting} & \textbf{FSR} & \textbf{DCPA} & \textbf{S2PA} & \textbf{Dice} \\
\hline\hline
FedAvg                & & & & 71.14\,$\pm$\,1.10 \\
\hline
\multicolumn{5}{l}{\textit{\textbf{Individual contribution}}} \\
\rowcolor{mylightred}
\quad +\,FSR          & \ding{51} & & & 72.55\,$\pm$\,1.13 \\
\quad +\,DCPA         & & \ding{51} & & 72.21\,$\pm$\,0.43 \\
\rowcolor{mylightred}
\quad +\,S2PA         & & & \ding{51} & \underline{73.81\,$\pm$\,1.09} \\
\hline
\multicolumn{5}{l}{\textit{\textbf{Cumulative}}} \\
\quad FSR\,+\,DCPA    & \ding{51} & \ding{51} & & 73.20\,$\pm$\,1.28 \\
\rowcolor{oursrow}
\textbf{FedBCS+}      & \ding{51} & \ding{51} & \ding{51} & \textbf{75.47\,$\pm$\,0.86} \\
\hline\thickhline
\end{tabular}
\end{table}
Added individually to FedAvg, every module improves Dice: S2PA by 2.67
points, FSR by 1.41 points, and DCPA by 1.07 points.
Cumulatively, combining FSR and DCPA achieves 73.20\% Dice, and
adding S2PA raises performance to 75.47\%. The full model exceeds
each single-module variant by 1.66--3.26 Dice points, so no single
component accounts for the full gain. Table~\ref{tab:dcpa_analysis} further compares DCPA
with the coupled CDPA module from the conference version.

\subsubsection{Fine-grained Analysis of DCPA}
\label{sec:dcpa_fine}

To examine the role of each prototype type in DCPA, we compare alignment
using semantic (center) prototypes, structural (boundary) prototypes, and
both types jointly.
\begin{table}[t]\small
\caption{\textbf{Fine-grained analysis of DCPA on Histology.}
The alignment module is evaluated on a single cross-validation
split, with Dice scores expressed as percentages. We compare semantic-only,
structural-only, and joint alignment with the coupled CDPA module from the
conference version, marked by a dagger. The best result is shown in
\textbf{bold}.}
\label{tab:dcpa_analysis}
\vspace{-10pt}
\centering
\small
\setlength\tabcolsep{6pt}
\renewcommand\arraystretch{1.2}
\begin{tabular}{lIc}
\hline\thickhline
\rowcolor{mydarkgreen}
\textbf{Alignment target} & \textbf{Dice} \\
\hline\hline
\rowcolor{mylightgreen}
Semantic prototype only          & 70.87 \\
Structural prototype only        & 70.30 \\
\rowcolor{mylightgreen}
Coupled prototype (CDPA$^\dagger$) & 71.50 \\
\textbf{Joint (DCPA)}            & \textbf{72.23} \\
\hline\thickhline
\end{tabular}
\end{table}
Table~\ref{tab:dcpa_analysis} shows that semantic-only and structural-only
alignment achieve 70.87\% and 70.30\% Dice, respectively. Aligning both
types jointly achieves the highest score of 72.23\%, as the two types
capture complementary cues, namely regional information and contours.
Joint alignment also outperforms the coupled CDPA prototype (71.50\%), so
separate semantic and structural branches guide alignment better than a
shared prototype.

\subsubsection{Hyperparameter Sensitivity}
\label{sec:hyperparam}

\begin{figure}[t]
\centering
\includegraphics[width=0.32\columnwidth]{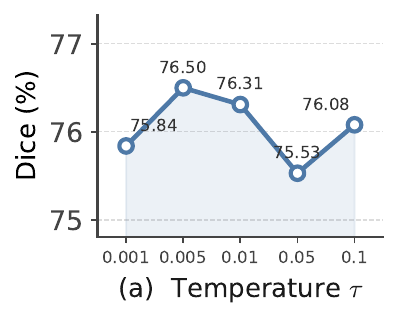}\hfill
\includegraphics[width=0.32\columnwidth]{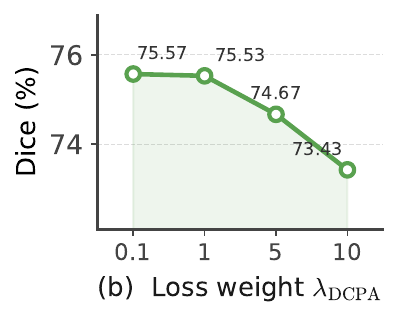}\hfill
\includegraphics[width=0.32\columnwidth]{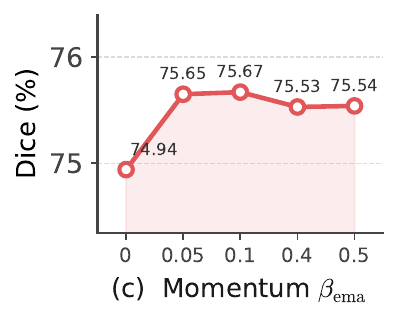}
\caption{\textbf{Hyperparameter sensitivity on Histology.}
Panels (a)--(c) show the effects of the DCPA contrastive temperature $\tau$,
the DCPA loss weight $\lambda_{\mathrm{DCPA}}$, and the S2PA aggregation
momentum $\beta_{\mathrm{ema}}$, respectively. Each parameter is varied
in a single cross-validation run while the remaining parameters are
fixed at their default values.}
\label{fig:hyperparam}
\end{figure}

We analyze sensitivity to three key hyperparameters on Histology
and report results in Fig.~\ref{fig:hyperparam}.

\noindent\textbf{(a)~Contrastive temperature $\tau$.}
The temperature controls the sharpness of the contrastive loss in DCPA.
Varying $\tau$ from 0.001 to 0.1, including one value below the selection
grid, yields Dice scores between 75.53\% and 76.50\%, a spread of less
than one Dice point. Because target
size and variability differ across benchmarks, we select $\tau$ separately
for each benchmark.

\noindent\textbf{(b)~DCPA loss weight $\lambda_{\mathrm{DCPA}}$.}
The default $\lambda_{\mathrm{DCPA}}=1.0$ balances
$\mathcal{L}_{\text{DCPA}}$ with $\mathcal{L}_{\text{dice}}$.
Performance remains stable as $\lambda_{\mathrm{DCPA}}$ varies from 0.1 to 1.
Larger weights impose excessive regularization on local training, reducing
Dice by approximately two points at $\lambda_{\mathrm{DCPA}}=10$.

\noindent\textbf{(c)~S2PA aggregation momentum $\beta_{\mathrm{ema}}$.}
This coefficient controls the contribution of the current prototype divergences to the
momentum update. Setting $\beta_{\mathrm{ema}}=0$ disables the adaptive update and
reduces S2PA to uniform parameter averaging, yielding the lowest performance
among the tested settings. All tested values of $\beta_{\mathrm{ema}}$ from 0.05 to 0.5 yield similar
Dice scores (75.53--75.67\%), and the default $\beta_{\mathrm{ema}}=0.4$ lies within
this stable region.

\subsection{Further Analysis}
\label{sec:further}

\subsubsection{Training Dynamics}

\begin{figure}[!t]
\centering
\begin{subfigure}[b]{0.5118\columnwidth}
\includegraphics[width=\linewidth]{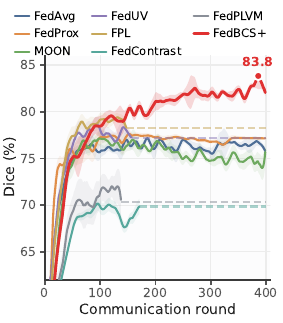}
\caption{Training dynamics}
\end{subfigure}\hfill
\begin{subfigure}[b]{0.4682\columnwidth}
\includegraphics[width=\linewidth]{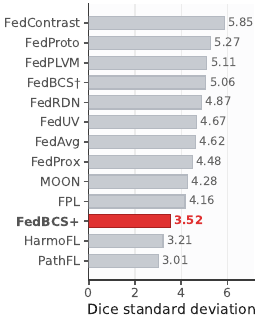}
\caption{Cross-client deviation}
\end{subfigure}
\caption{\textbf{Convergence and cross-client consistency on Prostate.}
(a)~Dice scores across communication rounds in a single
cross-validation run. Shaded bands indicate round-to-round fluctuations, and
dashed lines extend runs that stopped early at their final value. FedBCS+
reaches a peak Dice score of 83.8\% and remains the highest curve from
round 142 onward.
(b)~Cross-client standard deviation of the per-client mean Dice scores in
Table~\ref{tab:compare_sota} for all compared methods. Lower values indicate
more consistent performance. FedBCS+ reduces the standard deviation from 4.62
for FedAvg to 3.52; only PathFL and HarmoFL obtain lower values, both with
lower mean Dice.}
\label{fig:convergence}
\end{figure}

Fig.~\ref{fig:convergence} plots Dice against communication rounds in a
single cross-validation run on Prostate.

FedBCS+ differs from the other methods in two respects.
\textbf{(1)~Higher peak.}
FedBCS+ reaches 83.8\% Dice, compared with 80.4\% for FPL, 80.1\% for
FedUV, 78.7\% for FedProx, and at most 78.5\% for the remaining methods.
\textbf{(2)~Sustained improvement.}
FedPLVM, FPL, FedUV, and FedContrast stop early between rounds 138 and 171,
and FedAvg, FedProx, and MOON remain at about 75--77\% after round 200.
FedBCS+ stays above all other curves from round 142 onward and continues
to improve, from 79.0\% on average over rounds 100--150 to 82.1\% over
rounds 300--399.

\subsubsection{Frequency-domain Style Separation}
\label{sec:fsr_analysis}

\begin{figure*}[!t]
\centering
\begin{subfigure}[b]{0.22\linewidth}
\includegraphics[width=\linewidth]{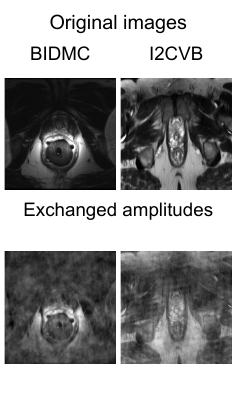}
\caption{Amplitude exchange}
\end{subfigure}
\hfill
\begin{subfigure}[b]{0.36\linewidth}
\includegraphics[width=\linewidth]{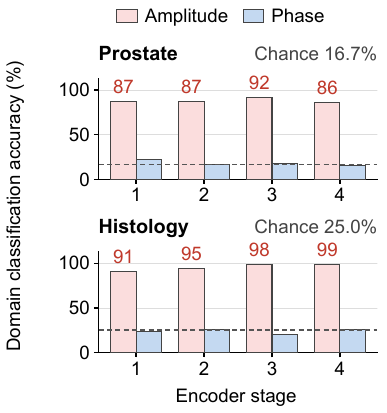}
\caption{Domain separability across stages}
\end{subfigure}
\hfill
\begin{subfigure}[b]{0.39\linewidth}
\includegraphics[width=\linewidth]{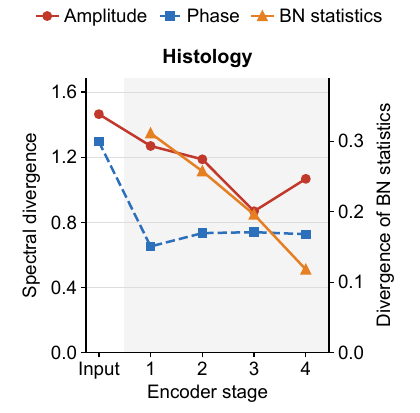}
\caption{Feature divergence across domains}
\end{subfigure}
\caption{\textbf{Frequency-domain style separation.}
(a)~Exchanging amplitude spectra between images from two Prostate clients
transfers appearance while preserving anatomy.
Each column retains the phase of its original image after the exchange.
(b)~Across encoder stages on Prostate and Histology, domain classification
accuracy reaches $86\text{--}99\%$ using amplitude spectra but remains near
chance using phase spectra.
(c)~On Histology, with a separate model trained for each domain, amplitude divergence
consistently exceeds phase divergence throughout the encoder. The left axis
shows spectral divergence, and the right axis shows divergence of batch
normalization statistics. These results demonstrate persistent style
discrepancies in intermediate features.}
\label{fig:fsr_analysis}
\end{figure*}

Fig.~\ref{fig:fsr_analysis} examines the frequency-domain separation of style
and content that motivates FSR (Sec.~\ref{sec:fsr}). In panel~(a), swapping
amplitude spectra between images from two Prostate clients transfers
intensity and contrast while preserving anatomical structure, illustrating
the distinct roles of amplitude and phase.
To quantify this separation in intermediate features, we train a linear
domain classifier on the encoder features at each stage of a FedAvg-trained
UNet. As shown in panel~(b), amplitude spectra provide strong domain
discrimination, with accuracies of $86\text{--}92\%$ on Prostate and
$91\text{--}99\%$ on Histology across all encoder stages. Classification
from phase spectra remains close to the respective chance levels of
$16.7\%$ on Prostate and $25\%$ on Histology.
Domain-specific appearance information thus persists beyond the input into
the amplitude spectra of intermediate features.
On Histology, we further train a separate model for each domain and measure
the cross-domain divergence of features at each layer (panel~(c)). Amplitude
divergence remains higher than phase divergence across all stages, with
amplitude-to-phase divergence ratios of approximately $1.1$ at the input and
$1.2\text{--}1.9$ at the feature stages. Since this pattern also holds
under independent training, style discrepancies need to be handled
throughout feature extraction.

\subsubsection{Prototype Entanglement Analysis}
\label{sec:dcpa_analysis}

\begin{figure}[!t]
\centering
\includegraphics[width=0.82\columnwidth]{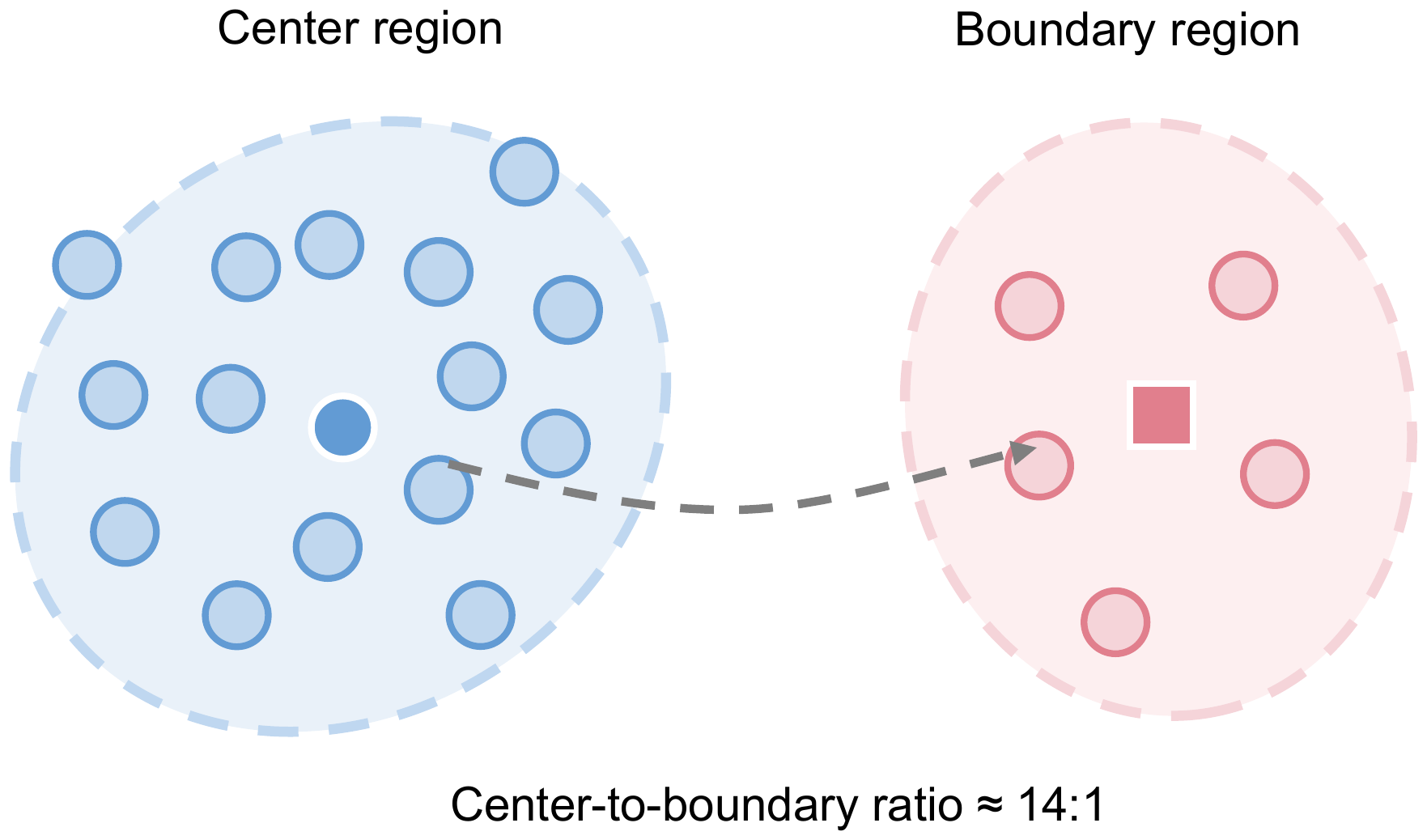}\\[5pt]
\includegraphics[width=0.99\columnwidth]{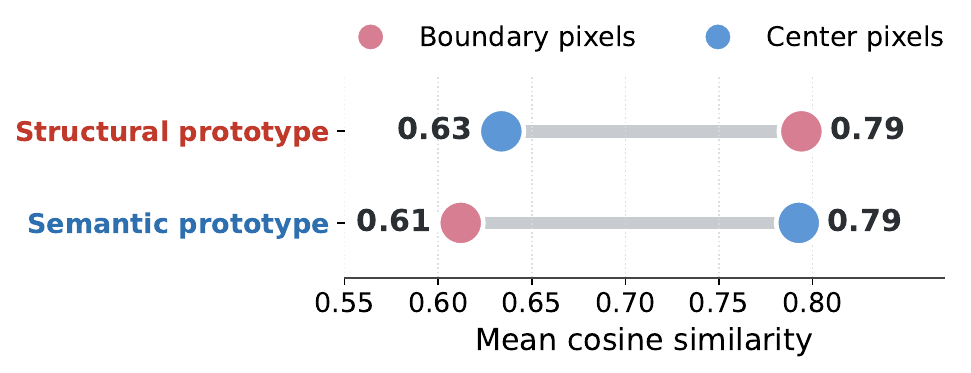}
\caption{\textbf{Prototype entanglement on Polyp.}
Prototypes are computed from fused decoder features. The upper panel shows
that center pixels are approximately 14 times as numerous as boundary
pixels, causing a shared foreground prototype to be dominated by center
features. The lower panel compares mean cosine similarities between each
prototype and boundary or center features. Structural prototypes align
more closely with boundary features, while semantic prototypes align more
closely with center features, motivating separate alignment.}
\label{fig:dcpa_coupled}
\end{figure}

Fig.~\ref{fig:dcpa_coupled} examines the representational coupling that
motivates DCPA (Sec.~\ref{sec:dcpa}). In the fused decoder stream on
Polyp, center pixels outnumber boundary pixels by approximately
$14{:}1$ (upper panel). As shown in the lower panel, each dedicated prototype is most
similar to the region it is built from: the structural prototype reaches
$0.79$ with boundary features and the semantic prototype $0.79$ with center
features, compared with $0.61$--$0.63$ for the opposite region. A coupled
prototype formed by averaging all foreground pixels, which is not shown in
the figure, is dominated by the center region: its cosine similarity is $0.79$ to
center features but only $0.63$ to boundary features.
The semantic and structural prototypes have a cosine similarity of $0.77$,
indicating that they are related but not interchangeable.
DCPA preserves this distinction by constructing and aligning the two
prototype types independently.

\subsubsection{Communication and Computation Overhead}
\label{sec:overhead}

\noindent\textbf{Communication.}
As summarized in Table~\ref{tab:method_analysis},
FedBCS+ uploads at most six prototypes per client per round on the binary
benchmarks: for each of the two fused streams (encoder and decoder), one
background and one foreground \emph{semantic} prototype, plus one foreground
\emph{structural} prototype. Compared with
FedBCS, which uploads four prototypes, DCPA adds one structural prototype per stream.
Each prototype contains 768 floating-point values, giving a maximum payload
of approximately 18\,KB. This amounts to 0.06\% of the 33\,MB of model
parameters transmitted each round. Prototype payloads remain in the kilobyte
range for the other compared methods, reaching hundreds of kilobytes for
FedPLVM with approximately 212 prototypes.

\noindent\textbf{Computation.}
On the shared UNet backbone (Table~\ref{tab:method_analysis}), FSR is
the only component that increases the client model size. It adds 902
parameters, increasing the total from 8,266,626 to 8,267,528, or by 0.01\%.
Each of the three FSR blocks contains a per-channel affine pair and two
mixing logits. For a $256 \times 256$ input, the FFT and inverse FFT increase
the computational cost of a forward pass from 24.147 to 24.383\,GFLOPs,
an increase of 0.98\%.
Beyond FedBCS, FedBCS+ introduces two additional operations
(Sec.~\ref{sec:discussion}): (i)~decoupled prototype construction in DCPA,
which requires one max-pooling pass for erosion and reuses existing features
without additional forward passes; and (ii)~prototype divergence computation
in S2PA, performed once on the server per round. Both costs are small; only
the server-side cost grows with the number of clients, which is at most nine
here.

\subsubsection{Theoretical Interpretation}

The convergence analysis in Sec.~\ref{sec:convergence} relates the proposed
components to the one-round objective bound in Theorem~1. With the default
$\lambda_{\mathrm{DCPA}}=1$, the effective second-moment term is
$\bar{\sigma}_{w,\rho_+}^2=\sigma_w^2+2\lambda_{\text{str}}\rho_+$, and
$C_{\mathrm{pa}}$ bounds the prototype-anchor perturbation.

\noindent\textbf{(1)~Prototype-guided aggregation.}
S2PA replaces sample-size-based aggregation with weights $\{w_m\}$ derived from
prototype divergences. As shown in Sec.~\ref{sec:convergence},
reducing the aggregation-dependent term $\sigma_w^2$ decreases the effective
second-moment term $\bar{\sigma}_{w,\rho_+}^2$ and tightens the one-round
descent bound; with round-varying weights, the stationarity bound holds up
to the weight-drift term $\Xi_R$ of Appendix~E. The analysis thus
motivates adaptive weighting, and the empirical results agree with it: FedBCS+ reaches a higher plateau in
Fig.~\ref{fig:convergence} and lowers $\sigma_{\mathrm{Dice}}$ relative to
FedAvg on all four 2D benchmarks.

\noindent\textbf{(2)~Decoupled semantic--structural alignment.}
DCPA separates the two prototype types to address the interaction introduced
by coupled alignment. The analysis captures this interaction through
$\rho_+$: a reduction from $\rho_+^{\text{CDPA}}$ to
$\rho_+^{\text{DCPA}}$ reduces the effective second-moment term
$\bar{\sigma}_{w,\rho_+}^2$. The standalone decoupling results in
Sec.~\ref{sec:dcpa_fine} are consistent with this analysis.

\section{Conclusion}\label{sec:conclusion}

We have presented FedBCS+, a federated learning framework for heterogeneous medical image segmentation that addresses incomplete contextual representations and layerwise style and aggregation biases. FSR recalibrates the amplitude spectra of intermediate features to obtain style-purified prototypes, DCPA constructs separate semantic and structural prototypes from multi-level features and aligns them independently, and S2PA converts each client's prototype divergence from the global consensus into adaptive aggregation weights. Experiments on five heterogeneous benchmarks spanning histopathology, MRI, ultrasound, and colonoscopy show that FedBCS+ achieves the highest mean Dice among the compared methods, with gains of up to 4.33 Dice points over FedAvg, and reduces cross-client performance variation relative to FedAvg on all four 2D benchmarks. The convergence analysis further characterizes how prototype-guided aggregation and decoupled alignment enter the optimization bound. The current framework assumes a shared network architecture and has been evaluated with up to nine clients; future work will extend prototype-guided aggregation to heterogeneous architectures, larger federations, and partial client participation.

\ifCLASSOPTIONcaptionsoff
  \newpage
\fi

\vspace{-4pt}
{\small
\bibliographystyle{IEEEtran}
\bibliography{ref}
}

\end{document}